\documentclass[10pt,twocolumn,letterpaper]{article}

\usepackage[pagenumbers]{cvpr} % To force page numbers, e.g. for an arXiv version

\usepackage{amsmath,amsfonts,bm}

\def\eqref#1{equation~\ref{#1}}
\def\1{\bm{1}}

\DeclareMathAlphabet{\mathsfit}{\encodingdefault}{\sfdefault}{m}{sl}
\SetMathAlphabet{\mathsfit}{bold}{\encodingdefault}{\sfdefault}{bx}{n}

\usepackage{times}

\usepackage{graphicx}
\usepackage{amsmath}
\usepackage{amssymb}
\usepackage{booktabs}
\usepackage{epsfig}
\usepackage{algorithm}
\usepackage{algorithmic}

\usepackage{url}
\usepackage{multirow}
\usepackage{diagbox}
\usepackage{tabularx}
\usepackage{caption}
\usepackage{graphicx}
\usepackage{subcaption}
\usepackage{adjustbox}
\usepackage{enumitem}
\usepackage{soul}
\usepackage{slashbox}

\usepackage{xhfill}
\usepackage{subfiles}

\newcommand{\be}{\begin{equation}}
\newcommand{\ee}{\end{equation}}

\usepackage[pagebackref=true,breaklinks=true,colorlinks,bookmarks=false]{hyperref}
\usepackage[bookmarks=false]{hyperref}

\usepackage[capitalize]{cleveref}
\crefname{section}{Sec.}{Secs.}
\Crefname{section}{Section}{Sections}
\Crefname{table}{Table}{Tables}
\crefname{table}{Tab.}{Tabs.}

\def\cvprPaperID{3846} % *** Enter the CVPR Paper ID here
\def\confName{CVPR}
\def\confYear{2022}

\begin{document}

%\title{Robustness Benchmarking with 3D Informed Corruptions}
%\title{Evaluating and Improving Robustness Using 3D Distribution Shifts}

\title{3D Common Corruptions and Data Augmentation}

\author{
O\u{g}uzhan Fatih Kar
\quad\quad
Teresa Yeo
\quad\quad
Andrei Atanov
\quad\quad
Amir Zamir
\vspace{3pt} \\ \vspace{10pt}
~~Swiss Federal Institute of Technology (EPFL) \\
\small\url{https://3dcommoncorruptions.epfl.ch/}
}

\maketitle

% \twocolumn [{%
%         \renewcommand\twocolumn[1][]{#1}%
%         \maketitle
%         \centering
%         \vspace{-5mm}
%         % \vspace{-97mm}
%         \hspace{1mm}
%         \includegraphics[scale=0.40]{images/fig1_pull.PNG}\vspace{-0pt}
%         % \vspace{2mm}
%         \captionof{figure}{\footnotesize{ \textbf{TODO, too long} We propose several new corruptions for RGB images informed by 3D structure of real world AND the corresponding labels for several dense prediction tasks such as surface normals and depth. Left, we apply several 3D informed corruptions like 3D defocusing, volumetric fog, or flash to Taskonomy images. Right, we render images with smooth trajectories, view jitter, changing field of view, camera roll and pitch. We show these corruptions are challenging for state of the art robustness methods. The proposed corruptions also enable developing new data augmentation techniques improving model robustness in real corrupted data.
%         % \vspace{8mm}
%         }}%
%         \vspace{8mm}
%         \label{fig:pull}
%     }]

%%%%%%%%% ABSTRACT
\begin{abstract}
%   We introduce a set of 3D input transformations enabling new corruptions. The distinction is that the corruptions exhibit practical changes in the RGB image such as depth of field, lighting, camera parameters, occlusions, and more. Thus, they are more likely to happen in the real world in contrast to using only 2D transformations. 
   
%   \textbf{TODO:Update} 
   %We introduce a set of input transformations enabling the generation of new corruptions on real images to analyze model robustness. The distinction is that the generated corruptions leverage scene 3D, thus they are more likely to happen in the real world compared to using only 2D transformations. 
%   In contrast to using only 2D transformations, our corruptions are more likely to happen in the real world. 
   %The transformations are efficient to compute and are controllable as we generate them programmatically with exposed parameters. Furthermore, they are extendable to standard vision datasets that do not readily come with 3D information. We find that robustness issues exposed by the proposed corruptions strongly correlate with realistic corruptions.
%   the performance of the methods aiming to improve robustness reduce drastically under these corruptions.
   %Motivated by this, we also introduce new 3D data augmentations that significantly improve robustness to real world distribution shifts compared to using only 2D augmentations. 
   
   We introduce a set of image transformations that can be used as \textbf{corruptions} to evaluate the robustness of models as well as \textbf{data augmentation} mechanisms for training neural networks. The primary distinction of the proposed transformations is that, unlike existing approaches such as Common Corruptions~\cite{hendrycks2019benchmarking}, the geometry of the scene is incorporated in the transformations -- thus leading to corruptions that are more likely to occur in the real world. 
   We also introduce a set of \textbf{semantic} corruptions (e.g. natural object occlusions. See Fig.~\ref{fig:compare2d}). 
   
   %Examples include shallow depth of field, changes in lighting and camera parameters, natural occlusions, and more.
   We show these transformations are `efficient' (can be computed on-the-fly), `extendable' (can be applied on most image datasets), expose vulnerability of existing models, and can effectively make models more robust when employed as `3D data augmentation' mechanisms. The evaluations on several tasks and datasets suggest incorporating 3D information into benchmarking and training opens up a promising direction for robustness research.

\end{abstract}

%%%%%%%%% BODY TEXT

% \textbf{TODO: is ``new 3D data augmentations" confusing?}
\section{Introduction}
\label{sec:intro}

%For pull fig: We propose several new corruptions for RGB images informed by 3D structure of real world AND the corresponding labels for several dense prediction tasks such as surface normals and depth. Left, we apply several 3D informed corruptions like 3D defocusing, volumetric fog, or flash to Taskonomy images. Right, we render images with smooth trajectories, view jitter, changing field of view, camera roll and pitch. We show these corruptions are challenging for state of the art robustness methods. The proposed corruptions also enable developing new data augmentation techniques improving model robustness in real corrupted data.

%Fig curves: We observe 2D augs can improve performance for \textit{some} 3D ones such as defocus or motion blur. See supplementary for details. However it doesnt help with others like fog or shadows.

% Neural networks deployed in the real world will encounter data with naturally occurring distortions, e.g. motion blur, or adversarial ones. Such changes make up shifts from the training data distribution. While neural networks are able to learn complex functions in-distribution, their predictions are profoundly unreliable under such shifts. This presents a core challenge that needs to be solved for these models to be useful in the real world.

\begin{figure}[!ht]
\centering
  \includegraphics[scale=0.064]{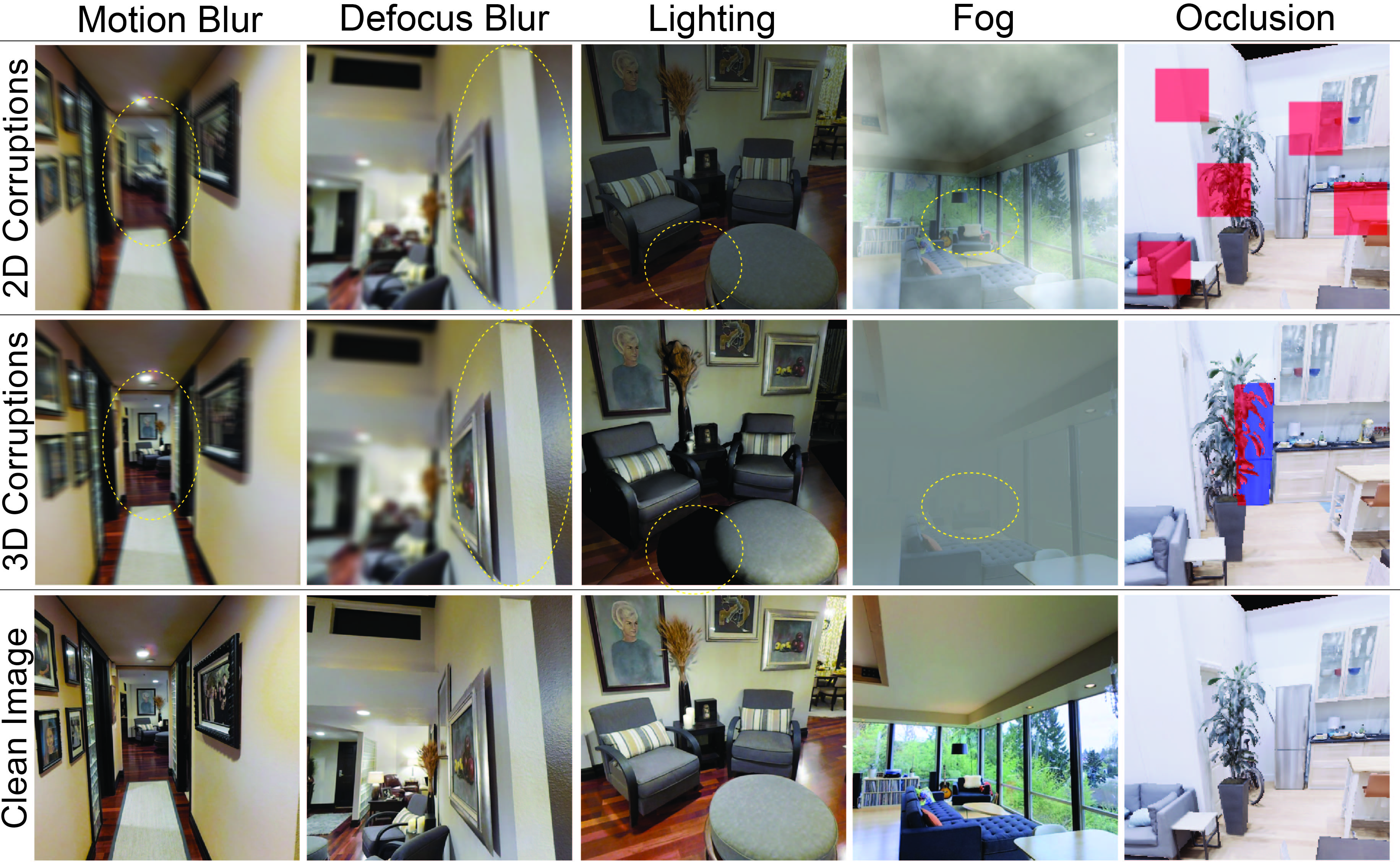}\vspace{-0pt}
%   Key differences between 3D CC and 2D CC.
% {\color{teal} \textbf{TODO:} update figure with amir's comments. add random cuts for 2d occlusion. add clean row.} 
\caption{\footnotesize{\textbf{Using 3D information to generate real-world corruptions.} The top row shows sample 2D corruptions applied uniformly over the image, e.g. as in Common Corruptions~\cite{hendrycks2019benchmarking}, disregarding 3D information. This leads to corruptions that are unlikely to happen in the real world, e.g. having the same motion blur over the entire image irrespective of the distance to camera~(top left). Middle row shows their 3D counterparts from 3D Common Corruptions~(3DCC). The circled regions highlight the effect of incorporating 3D information.
% the selected corruptions from 2D common corruptions (2DCC) and the bottom their 3D counterpart (3DCC). 3DCC incorporates characteristics of realistic corruptions while 2DCC applies the corruption \textit{uniformly} over the image, \textit{ignoring 3D information}. 
More specifically, in 3DCC, \textbf{1.} \textbf{motion blur} has a \textit{motion parallax} effect where objects further away from the camera seem to move less, \textbf{2.} \textbf{defocus blur} has a \textit{depth of field} effect, akin to a large aperture effect in real cameras, where certain regions of the image can be selected to be in focus,  \textbf{3.} \textbf{lighting} takes the scene geometry into account when illuminating the scene and casts shadows on objects, \textbf{4.} \textbf{fog} gets denser further away from the camera, \textbf{5. occlusions} of a target object, e.g. fridge~(blue mask), are created by changing the camera's viewpoint and having its view \textit{naturally obscured by another object}, e.g. the plant~(red mask). This is in contrast to its 2D counterpart that randomly discards patches~\cite{devries2017improved}. See \href{https://3dcommoncorruptions.epfl.ch}{project page} for a video version of the figure.}}\label{fig:compare2d}\vspace{-12pt}
\end{figure}

\begin{figure*}[!ht]
\centering
  \includegraphics[width=0.95\textwidth]{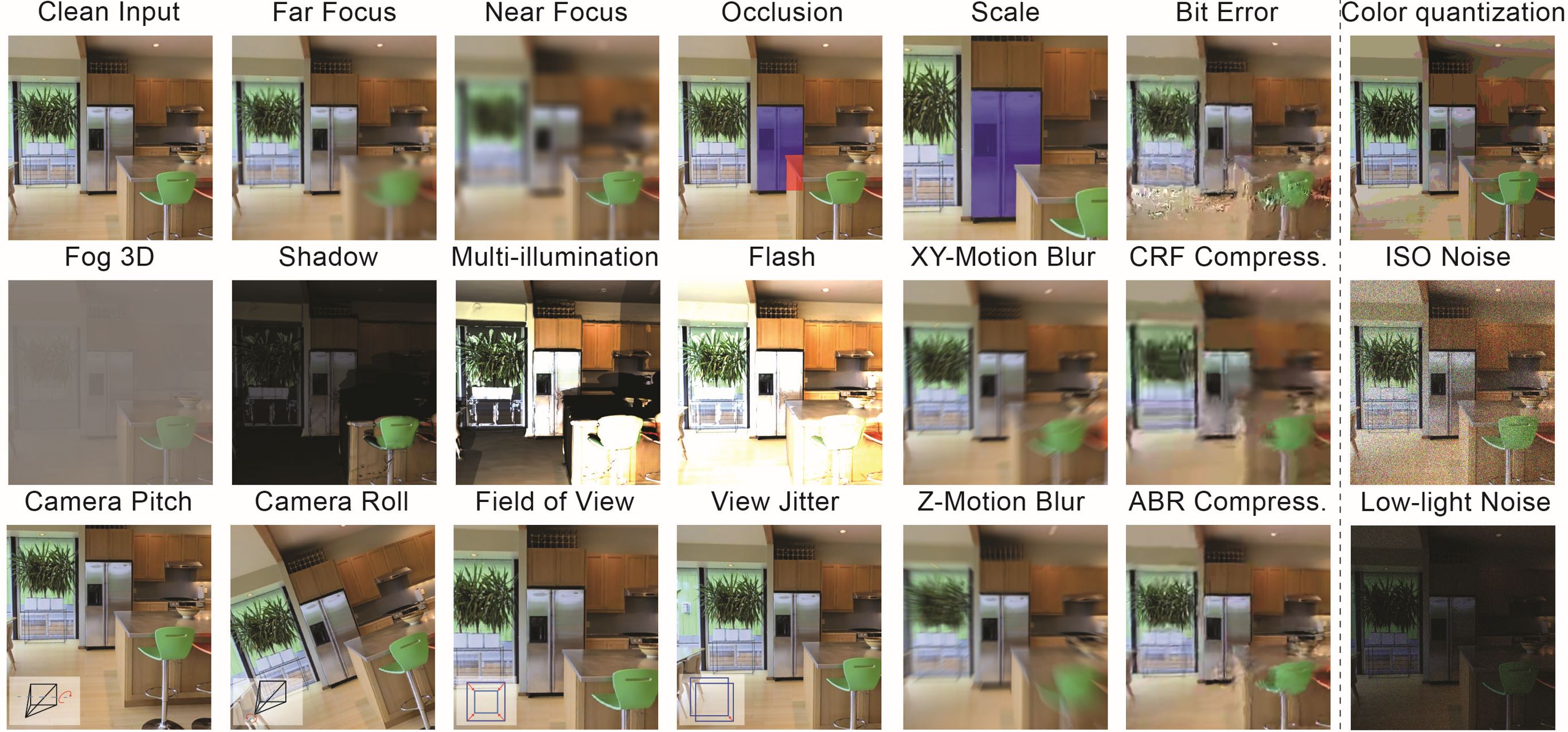} %\vspace{-0pt}
\caption{\footnotesize{
% {\color{blue} \textbf{Amir}, re your comment about \textit{"new", since not all are 3D}, we assumed in the entire paper that all corruptions are called 3DCC. Unless we find another name for all corruption types, its easier to use 3DCC.}\\
% {\color{red}\textbf{TODO}: Amir: I agree. I'm not suggesting to call things smth other than 3DCC. But it's undeniable and impossible to miss that some of these corruptions have nothing to do with 3D, so we have to clarify that somewhere and this plot is a good opportunity. If we dont wear that on our sleeves, the reviewers will use it as a "catch" and criticize. Without having to change the name, you can give the figure an organization simiar to what I suggested and use the caption to clarify that ``A subset of the corruptions marked in the right two columns are novel and commonly faced in the real world, but are not 3D based. We include them in our benchmark." }\\
% {\color{teal}\textbf{TODO}: update figure with amir's comments}\\
\textbf{The new corruptions.} We propose a \textit{diverse} set of new corruption operations ranging from defocusing (near/far focus) to lighting changes and 3D-semantic ones, e.g. object occlusion. These corruptions are all \textit{automatically generated}, \textit{efficient} to compute, and can be applied to \textit{most datasets}~(Sec.~\ref{sec:3dccstandard}). We show that they expose vulnerabilities in models~(Sec.~\ref{sec:3dcc_issues}) and are a good approximation of {realistic corruptions}~(Sec.~\ref{sec:3dvsreal}). A subset of the corruptions marked in the last column are novel and commonly faced in the real world, but are not 3D based. We include them in our benchmark. For occlusion and scale corruptions, the blue and red masks denote the amodal visible and occluded parts of an object, e.g. the fridge.}}\label{fig:pull} \vspace{-15pt}
\end{figure*}

% and can be used as augmentations to improve model robustness to real world corruptions (see Sec.~\ref{sec:3daugs_exp}).

% We propose several new corruptions for RGB images informed by 3D structure of real world AND the corresponding labels for several dense prediction tasks such as surface normals and depth. Left, we apply several 3D informed corruptions like 3D defocusing, volumetric fog, or flash to Taskonomy images. Right, we render images with smooth trajectories, view jitter, changing field of view, camera roll and pitch. We show these corruptions are challenging for state of the art robustness methods. The proposed corruptions also enable developing new data augmentation techniques improving model robustness in real corrupted data.

Computer vision models deployed in the real world will encounter naturally occurring distribution shifts from their training data. These shifts range from lower-level distortions, such as motion blur and illumination changes, to semantic ones, like object occlusion. 
% \ty{These shifts range from image corruptions such as motion blur to changes in camera extrinsics, or semantic ones like occlusions. (removed illumination because its sounds like illumination and camera extrinsics are in the same "category".)}
Each of them represents a possible failure mode of a model and has been frequently shown to result in profoundly unreliable predictions~\cite{dodge2017study,hendrycks2019benchmarking,szegedy2013intriguing,jo2017measuring,geirhos2020shortcut}. Thus, a systematic testing of vulnerabilities to these shifts is critical before deploying these models in the real world.

%Thus, it is critical to systematically identify model failure against a wide range of realistic shifts to reveal model vulnerabilities before they are deployed in the real world. %possible shifts that can happen in the real world.  
%, camera parameter changes e.g. FoV, to semantic shifts e.g. occlusions. Each of these shifts represent a possible failure mode of the model, which can result in profoundly unreliable predictions. We introduce a framework and dataset that can systematically identify a wide range of failure mode of any model. Thus, revealing their vulnerabilities before they are deployed in the real world.

% This work presents a framework to test model failures against distribution shifts. 
This work presents a set of distribution shifts in order to test models' robustness.
% The distinction is that we leverage 3D information to generate the shifts rather than performing uniform 2D modifications over the image~(See Fig.~\ref{fig:compare2d}).
In contrast to previously proposed shifts which perform uniform 2D modifications over the image, such as Common Corruptions~(2DCC)~\cite{hendrycks2019benchmarking}, our shifts incorporate 3D information to generate corruptions that are consistent with the scene geometry. This leads to shifts that are more likely to occur in the real world~(See Fig.~\ref{fig:compare2d}). The resulting set includes 20 corruptions, each representing a distribution shift from training data, which we denote as \textit{3D Common Corruptions}~(3DCC). 3DCC addresses several aspects of the real world, such as camera motion, weather, occlusions, depth of field, and lighting. Figure~\ref{fig:pull} provides an overview of all corruptions. As shown in Fig.~\ref{fig:compare2d}, the corruptions in 3DCC are more diverse and realistic compared to 2D-only approaches. 

% the resulting shifts are both diverse and more realistic compared to their 2D counterparts.

We show in Sec.~\ref{sec-exp} that the performance of the methods aiming to improve robustness, including those with diverse data augmentation, reduce drastically under 3DCC. Furthermore, we observe that the robustness issues exposed by 3DCC well correlate with corruptions generated via {photorealistic synthesis}. Thus, 3DCC can serve as a challenging testbed for real-world corruptions, especially those that depend on scene geometry.

Motivated by this, our framework also introduces new \textit{3D data augmentations}. They take the scene geometry into account, as opposed to 2D augmentations, thus enabling models to build invariances against more realistic corruptions. We show in Sec.~\ref{sec:3daugs_exp} that they significantly boost model robustness against such corruptions, including the ones that cannot be addressed by the 2D augmentations. 
% Thus, they offer a promising tool  improve robustness to distribution shifts, including 3D ones that can't be sufficiently addressed by 2D augmentations. Furthermore, they can be combined with 

% keeps track of the performance on real corruptions, 

% hence it can serve as a challenging and 

% can keep track of the performance on real corruptions. These indicate that 3DCC can serve as a challenging and useful test bed to assess model robustness. Motivated by this, we also introduce new \textit{3D augmentations} that significantly improves robustness to real world distribution shifts compared to several baselines. 

% Our framework to generate 3DCC and 3D data augmentations are \textit{controllable}, i.e. they are generated programmatically with exposed parameters, which can be used for fine-grained analysis of robustness to corruptions, e.g. increasing fog density. 

The proposed corruptions are \textit{generated programmatically} with \textit{exposed parameters}, enabling fine-grained analysis of robustness, e.g. by continuously increasing the 3D motion blur. 
They are \textit{efficient} to compute and can be computed on-the-fly during training as data augmentation with a small increase in computational cost.
% \ty{with negligible increase in computational cost?}.
They are also \textit{extendable}, i.e. they can be applied to standard vision datasets, e.g. ImageNet~\cite{deng2009imagenet}, that do not come with 3D labels. 

\section{Related Work}

% \ty{Importance of data evidenced in new subjournal\footnote{http://datasets.mlr.press/}, datasets track\footnote{https://neurips.cc/Conferences/2021/CallForDatasetsBenchmarks}}

% %This work has connections to a number of topics, including ensembling, uncertainty estimation and calibration, 
% Should this be in intro instead? 
% This work presents a data-focused approach to robustness. The critical role of data in robustness has been recently discussed in depth in a number of works~\cite{sambasivan2021everyone, paullada2020data,hendrycks2021unsolved}.

% “The critical role of data in robustness has been recently discussed in depth in a number of works x y”.

% \ty{Our work has connections to several topics e.g. those that aim to diagnose failures of neural networks via robustness benchmarks, photorealistic data generation techiniques and methods to improve robustness. We give an overview of some of them within the constraints of space.}

%This work presents a data-focused approach to robustness. The critical role of data in robustness has been recently discussed in detail in a number of works~\cite{sambasivan2021everyone, paullada2020data,hendrycks2021unsolved}. 

% Our work has connections to several topics, including robustness benchmarking and analysis, photorealistic data generation, and image restoratio

% e.g. those that aim to diagnose failures of models via robustness benchmarks, photorealistic data generation techniques and methods to improve robustness. We give an overview of some of them within the constraints of space.

This work presents a data-focused approach~\cite{sambasivan2021everyone, paullada2020data} to robustness. 
% and has connections to several topics. 
We give an overview of some of the related topics within the constraints of space.

\textbf{Robustness benchmarks based on corruptions:} Several studies have proposed robustness benchmarks to understand the vulnerability of models to corruptions. %These benchmarks differ in how realistic the corruptions are which corresponds to how expensive the data collection procedure is. 
%These benchmarks make a trade-off between realism and scalability of the generated corruptions. 
A popular benchmark, Common Corruptions (2DCC)~\cite{hendrycks2019benchmarking}, generates synthetic corruptions on real images that expose sensitivities of image recognition models. %under different categories: noise, blur, weather, and digital on real images. %Thus, it falls in the middle of the spectrum. 
% It makes use of 2D methods such as Gaussian kernels to efficiently generate corruptions. 
%from 2DCC do not use 3D information i.e. motion blur is applied uniformly over the image which does not correspond to motion blur in the real world. 
It led to a series of works either creating new corruptions or applying similar corruptions on other datasets for different tasks~\cite{kamann2020benchmarking,michaelis2019benchmarking,chattopadhyay2021robustnav,yi2021benchmarking,mintun2021interaction, sun2022benchmarking}. In contrast to these works, 3DCC modifies real images \textit{using 3D information} to generate realistic corruptions. The resulting images are both perceptually different and expose different failure modes in model predictions compared to their 2D counterparts~(See Fig.~\ref{fig:compare2d} and~\ref{fig:compare2d3dcorrs}). %makes use of 3D to generate corruptions while still applying on real images.
Other works create and capture the corruptions in the real world, e.g. ObjectNet~\cite{barbu2019objectnet}. Although being realistic, it requires significant manual effort and is not extendable. A more scalable approach is to use computer graphics based 3D simulators to generate corrupted data~\cite{leclerc20213db} which can lead to generalization concerns. 3DCC aims to generate corruptions \textit{as close to the real world} as possible while staying \textit{scalable}.

\textbf{Robustness analysis} works use \textit{existing} benchmarks to probe the robustness of different methods, e.g. data augmentation or self-supervised training, under several distribution shifts. Recent works investigated the relation between synthetic and natural distribution shifts~\cite{taori2020measuring, hendrycks2021many, miller2021accuracy, djolonga2021robustness} and effectiveness of architectural advancements~\cite{bhojanapalli2021understanding, shao2021adversarial, naseer2021intriguing}. We select several popular methods to show that 3DCC can serve as a challenging benchmark~(Fig.~\ref{fig:comparemodels} and ~\ref{fig:3dcc_issues}).
\begin{figure*}[!ht]
\centering
  \includegraphics[scale=0.065]{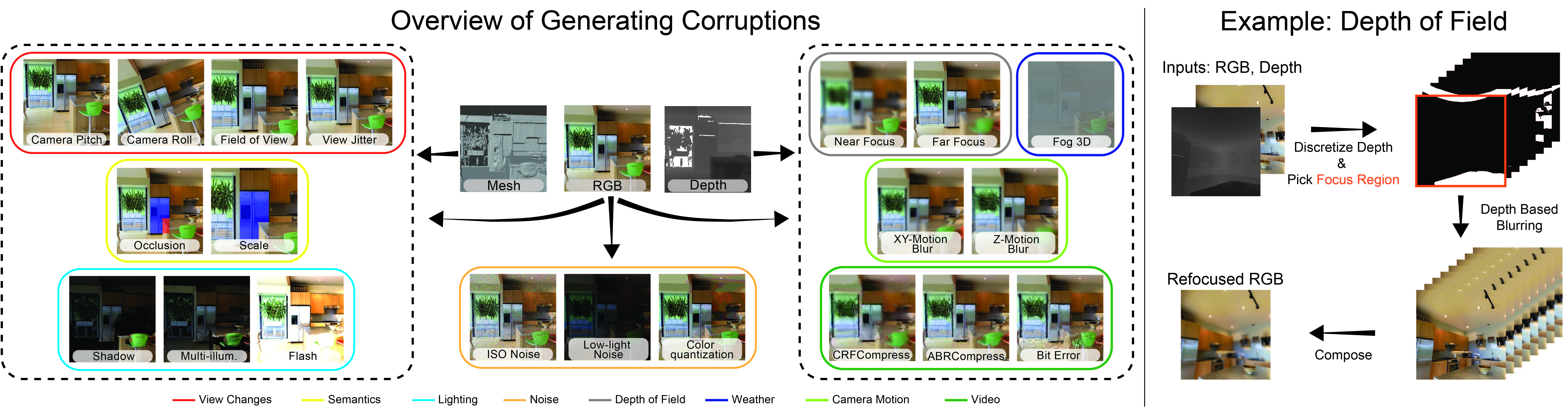}\vspace{-8pt}
%   How do we create 3D shifts? For all the corruptions on the right and bottom left, having RGB and scene depth is sufficient.
\caption{\textbf{Left:} We show the \textit{inputs} needed to create each of our corruptions, e.g. the 3D information such as depth, and RGB image. These corruptions have also been grouped (in solid colored lines) according to their \textit{corruption types}. 
% Thus, to create the distortions in the dashed box, one only needs the RGB image and their corresponding depth.
For example, to create the distortions in the dashed box in the right, one only needs the RGB image and its corresponding depth. For the ones in the left dashed box, 3D mesh is required. Note that one can create \textit{view changes} corruptions also from panoramic images if available, without a mesh.
% Given these inputs, creating e.g. near and far focus uses the depth of field method. We show in  Sec.~\ref{sec:3dccstandard} that these corruptions can be applied to any dataset, even those that do not come with depth (3D information). 
\textbf{Right:} As an example, we show an overview of generating depth of field effect \textit{efficiently}. The scene is first split into multiple layers by discretizing scene depth. Next, a region is chosen to be kept in focus~(here it is the region closest to the camera). We then compute the corresponding blur levels for each layer according to their distance from the focus region, using a pinhole camera model. The final refocused image is obtained by compositing blurred image layers. 
% \ty{\textbf{(Right)} (dont know how to explain properly) We show how we create the depth of field effect. 1. We first discretize depth into n bins. The scene, thus the depth, is split into these n regions. We then choose a region to keep in focus. The region with this depth range in the RGB image will be in focus, while the rest will be out of focus. In this example, the region closest to the camera is chosen to be in focus. 2. n RGB images are created according to these regions. Images further from the focus region gets blurred with higher intensity. 3. The images are composed to return, in this case, a near focused rgb.} 
}\label{fig:how}\vspace{-15pt}
\end{figure*}

\textbf{Improving robustness:} 
%\textit{Data augmentation:}(REWRITE, UPDATE!) One approach to addressing robustness involves using data augmentation during training~\cite{madry2017towards,zhang2017mixup,hendrycks2019augmix,kim2020learning,ashukha2020pitfalls,hendrycks2021many}. Such methods usually involve \emph{training with a set of corruptions} to generalize to unseen ones~\cite{rusak2020simple}. However, performance gains can be non-uniform, e.g. Gaussian noise augmentation improves performance on other noise corruptions (e.g. impulse, shot noise) but hurts performance on fog and contrast~\cite{ford2019adversarial}. Instead, our main mechanism uses a large set of middle domains (not corruptions) to be resistant to a wide range of diverse \textit{unseen} corruptions. We do not use any corruptions during training, except to calibrate uncertainty. Other robustness strategies includes ensembles and consistency constraints.
Numerous methods have been proposed to improve model robustness such as data augmentation with corrupted data~\cite{ford2019adversarial, madry2017towards,lopes2019improving,rusak2020increasing}, texture changes~\cite{geirhos2018imagenet,hendrycks2021many}, image compositions~\cite{zhang2017mixup,yun2019cutmix} and transformations~\cite{hendrycks2019augmix,yin2019fourier}. While these methods can generalize to some unseen examples, performance gains are non-uniform~\cite{rusak2020simple,ford2019adversarial}. Other methods include self-training~\cite{xie2020self}, pre-training~\cite{hendrycks2019using, orhan2019robustness}, architectural changes~\cite{bhojanapalli2021understanding,shao2021adversarial}, and diverse ensembling~\cite{pang2019improving,kariyappa2019improving,yang2020dverge,Yeo_2021_ICCV}. Here we instead adopt a data-focused approach to robustness by \textbf{i.} providing a large set of realistic distribution shifts and \textbf{ii.} introducing {new 3D data augmentation} that improves robustness against real-world corruptions~(Sec.~\ref{sec:3daugs_exp}).

\textbf{Photorealistic image synthesis} involves techniques to generate realistic images. Some of these techniques have been recently used to create corruption data. These techniques are generally specific to a single real-world corruption. Examples include adverse weather conditions~\cite{fattal2008single, sakaridis2018semantic, von2019simulating,hu2019depth, tremblay2021rain }, motion blur~\cite{brooks2019learning, niklaus20193d}, depth of field~\cite{potmesil1981lens, wang2018deeplens,wadhwa2018synthetic, eftekhar2021omnidata, barsky2008algorithms}, lighting~\cite{xu2018deep, helou2020vidit}, and noise~\cite{foi2008practical, wei2020physics}. They may be used for purely artistic purposes or to create training data. Some of our 3D transformations are instantiations of these methods, with the downstream goal of testing and improving model robustness in a unified framework with a wide set of corruptions.

%  \ty{\textbf{Photorealistic data} generation involves techniques to create realistic synthetic images. These techniques tend to be specific to a single real world corruption. Some examples of realisitc corruptions include adverse weather conditions~\cite{fattal2008single, sakaridis2018semantic, von2019simulating,hu2019depth, tremblay2021rain }, motion blur~\cite{brooks2019learning, niklaus20193d}, depth of field~\cite{potmesil1981lens, wang2018deeplens,wadhwa2018synthetic, eftekhar2021omnidata, barsky2008algorithms}, lighting~\cite{xu2018deep, helou2020vidit}, and noise~\cite{foi2008practical, wei2020physics}. They may be used for purely artistic purposes or to create training data. (I think there shouldnt be benchmarking, it sounds confusing). Some of our 3D transformations are instantiations of these methods, with the downstream goal of evaluating and improving robustness of neural networks.} 

\textbf{Image restoration} 
% All those realblur\cite{rim2020real}, sidd noise\cite{abdelhamed2018high}, reds\cite{Nah_2021_CVPR}, lighting vidit~\cite{helou2020vidit}, mff\cite{zhang2020real},  (?) papers goes here. Unlike them who wants to go from the corrupted image to clean image, we corrupt the image by respecting the 3D structure of the real world for robustness evaluation/improvement.
aims to undo the corruption in the image using classical signal processing techniques~\cite{kundur1996blind,fergus2006removing,elad2006image,mairal2009non} or learning-based approaches~\cite{zhang2017beyond,chen2018learning,nah2017deep,zhang2017learning,rim2020real,Nah_2021_CVPR,abdelhamed2018high}. We differ from these works by generating corrupted data, rather than removing it, to use them for benchmarking or data augmentation. %Thus, our focus is making the model to be invariant to corruptions encountered in the real world, as opposed to being able to remove them as a pre-processing step before being passed into the model.
Thus, in the latter, we train with these corrupted data to encourage the model to be invariant to corruptions, as opposed to training the model to remove the corruptions as a pre-processing step.

% We differ from these works by generating corrupted data rather than removing it. We then use this data as additional training data to undo the effect of corruption on the target task. \ty{We differ from these works by generating corrupted data rather than removing it, for the purpose of benchmarking the robustness of models or using them as data augmentations. Our focus is on training the model to be invariant to corruptions encountered in the real world, as opposed to being able to remove them as a pre-processing step before being passed into the model.}

\textbf{Adversarial corruptions} add imperceptible \textit{worst-case} shifts to the input to fool a model~\cite{szegedy2013intriguing,kurakin2016adversarial,madry2017towards,croce2020robustbench}. Most of the failure cases of models in the real world are not the result of adversarial corruptions but rather \textit{naturally occurring distribution shifts}. Thus, our focus in this paper is to generate corruptions that are likely to occur in the real world.

% \ty{\textbf{Adversarial corruptions} add imperceptible worst case shifts to the input to fool a model~\cite{szegedy2013intriguing,kurakin2016adversarial,madry2017towards}. However, most of the failure cases of models in the real world are not the result of adversarial corruptions but rather naturally occurring distribution shifts e.g. motion blur. Thus, our aim in this paper is to generate corruptions that are likely to occur in the real world.}

% \textbf{SSL:} Andrei

\section{Generating 3D Common Corruptions}
\subsection{Corruption Types}\label{sec:methods}

We define different corruption types, namely \textit{depth of field}, \textit{camera motion}, \textit{lighting}, \textit{video}, \textit{weather}, \textit{view changes}, \textit{semantics}, and \textit{noise}, resulting in 20 corruptions in 3DCC. 
Most of the corruptions require an RGB image and scene depth, while some needs 3D mesh~(See Fig.~\ref{fig:how}).
% Figure~\ref{fig:how} provides an overview of generating them from the inputs. 
We use a set of methods leveraging 3D synthesis techniques or image formation models to generate different corruption types, as explained in more detail below. Further details are provided in the \href{https://3dcommoncorruptions.epfl.ch/3DCC_supp.pdf}{supplementary}.

% We explain each in more detail below.
% Below we explain our methods to generate corruptions in more detail.

% {\color{teal}\textbf{TODO}: should corruption type names be changed e.g. depth of field - refocus blur, camera motion - camera motion blur, lighting - lighting changes, video - video compression, weather - adverse weather}\\
% {\color{teal}\textbf{TODO}: add significance of corruptions (per amir's comments)}\\
% {\color{teal}\textbf{TODO}: Near/far focus corruption was called defocusing previously. Choose from DoF, defocus or refocus?}\\

\noindent\textbf{Depth of field} corruptions create refocused images. They keep a part of the image in focus while blurring the rest.
% changes the focused region in the image generates images with different focus regions, while blurring the rest of the image.
We consider a layered approach~\cite{eftekhar2021omnidata, barsky2008algorithms} that splits the scene into multiple layers. For each layer, the corresponding blur level is computed using the pinhole camera model. The blurred layers are then composited with alpha blending. Figure~\ref{fig:how} (right) shows an overview of the process. We generate \textit{near focus} and \textit{far focus} corruptions by randomly changing the focus region to the near or far part of the scene. 
% Further details are given in the supplementary. {\color{brown} details for all are given in sup mat right?}

% \begin{figure}[!ht]
% \centering
%   \includegraphics[scale=0.40]{example-image-a}\vspace{-0pt}
% \caption{\footnotesize{ Generating depth of field }}\label{fig:defocus}
% \end{figure}

% We simulate depth of field effect by layering the scene into different depth ranges.

% Using the depth information we create a synthetic depth-of-field effect, similar to this,this,this, and focus to the near and far sides of the scene. We define near and far according to the hyperfocal distance, which is unique for each scene depending on the camera FoV and maximum depth. See the supplementary for details.

%  To generate, we first transform the input image into point cloud using the depth information.   Then,  we  define  a  random  camera  motion  and  render  novel  views  along  the  trajectory. As  the  point  cloud has partial information about the scene geometry, the rendered views will suffer from disocclusion artefacts.
 
\noindent\textbf{Camera motion} creates blurry images due to camera movement during exposure. To generate this effect, we first transform the input image into a point cloud using the depth information. Then, we define a trajectory (camera motion) and render novel views along this trajectory. As the point cloud was generated from a single RGB image, it has incomplete information about the scene when the camera moves. Thus, the rendered views will have disocclusion artifacts. To alleviate this, we apply an inpainting method from~\cite{niklaus20193d}. %In the final step, we perform averaging across views to generate motion blur that is parallax consistent.
The generated views are then combined to obtain parallax-consistent motion blur.
We define \textit{XY-motion blur} and \textit{Z-motion blur} when the main camera motion is along the image XY-plane or Z-axis, respectively. 

%\textbf{Camera Motion along XY and Z axes:} We model two main camera movements in XYZ plane. In the first one the move is dominantly along the planar XY plane with a small randomized change in the Z domain while it is reversed for the second case. We call them as XY-Motion Blur and Z-Motion Blur. 

% {\color{brown} \noindent\textbf{Lighting} corruptions change the scene illumination by either dimming the original light source or adding new ones. We use Blender~\cite{blender} to manipulate and compute the corresponding illumination for a given viewpoint in the 3D mesh. For the \textit{flash} corruption, a light source is placed on the camera, while for \textit{shadow} corruption it is placed in random diverse locations outside camera frustum. }

\noindent\textbf{Lighting} corruptions change scene illumination by adding new light sources and modifying the original illumination. We use Blender~\cite{blender} to place these new light sources and compute the corresponding illumination for a given viewpoint in the 3D mesh. For the \textit{flash} corruption, a light source is placed at the camera's location, while for \textit{shadow} corruption, it is placed at random diverse locations outside the camera frustum. % to generate shadows.
Likewise, for \textit{multi-illumination} corruption, we compute the illumination from a set of random light sources with different locations and luminosities.

% while for \textit{shadow} and \textit{multi-illumination} corruptions, we place them on diverse locations outside camera frustum. 
% The single light source is placed into the camera location to generate flash illumination~(leading to \textit{flash} corruption) and  
% for \textit{flash} corruption, and 
%Final image is obtained by multiplying the illumination with albedo. As we don't have material information, we approximate it using the RGB image. 
% \textbf{Flash:} We place a point light source to the camera location and compute the corresponding illumination in the scene which is effected from the scene depth and normals (cite taskonomy). Since we don't have albedo information, we treat original scene illumination as albedo and multiply with the lighting to obtain a flashed image.

% \textbf{Shadow and Lighting:} Similar to flash, but this time we placed several point sources in random locations to create diverse illuminations. For shadow we progressively select the relighted images with denser shadows with higher shift intensities. For lighting, we randomly select and add relighted images for each severity level and progressively reduced the number of relighted images used.

% Since we can create multiple frames to synthesize motion blur, we also leverage them to generate video corruptions for single images.

\noindent\textbf{Video} corruptions arise during the processing and streaming of videos. 
% To synthesize camera motion blur, we need to create multiple frames. These frames can also be used to generate video corruptions.
Using the scene 3D, we create a video using multiple frames \textit{from a single image} by defining a trajectory, similar to motion blur.
Inspired by~\cite{yi2021benchmarking}, we generate \textit{average bit rate (ABR)} and \textit{constant rate factor (CRF)} as H.265 codec compression artifacts, and \textit{bit error} to capture corruptions induced by imperfect video transmission channel. After applying the corruptions over the video, we pick a single frame as the final corrupted image.

% video compression and transmission artefacts as corruptions. Inspired by~\cite{yi2021benchmarking}, we generate \textit{H265 }

% For this we utilized ffmpeg inspired from~\cite{yi2021benchmarking}. We also included color quantization as another corruption which stems from using a reduced number of bits for the image. 
\noindent\textbf{Weather} corruptions degrade visibility by obscuring parts of the scene due to disturbances in the medium. We define a single corruption and denote it as \textit{fog 3D} to differentiate it from the fog corruption in 2DCC. We use the standard optical model for fog~\cite{fattal2008single,sakaridis2018semantic,von2019simulating}:
%that takes the scene depth into account:
\be
\mathbf{I(x)} ~=~ \mathbf{R(x)} \mathbf{t}(\mathbf{x}) + \mathbf{A}(1 - \mathbf{t}(\mathbf{x})) \label{fog},
\ee
where $\mathbf{I(x)}$ is the resulting foggy image at pixel $x$, $\mathbf{R(x)}$ is the clean image, $\mathbf{A}$ is atmospheric light, and $\mathbf{t}(\mathbf{x})$ is the transmission function describing the amount of light that reaches the camera. When the medium is homogeneous, the transmission depends on the distance from the camera, $\mathbf{t}(\mathbf{x})=\exp{(-\beta \mathbf{d}(\mathbf{x}))}$ where $\mathbf{d}(\mathbf{x})$  is the scene depth and $\beta$ is the attenuation coefficient controlling the fog thickness. 
% We employ Eq.~\ref{fog} and denote the resulting corruption as \textit{fog 3D} to differentiate it from the 2D fog corruption in 2DCC.

%We use the physics model of \cite{sakaridis2018semantic,von2019simulating} that takes the scene depth into account and intensifies the fog intensity accordingly.

\noindent\textbf{View changes} are due to variations in the camera extrinsics and focal length. Our framework enables rendering RGB images conditioned on several changes, such as \textit{field of view}, \textit{camera roll} and \textit{camera pitch}, using Blender. This enables us to analyze the sensitivity of models to various view changes in a controlled manner. We also generate images with \textit{view jitter} that can be used to analyze if models predictions flicker with slight changes in viewpoint. 
% to probe consistency among model predictions. 
%See Fig.~\ref{fig:camera} for an overview. %by conditioning RGB rendering on camera parameters. 
%such as continuously increasing \textit{field of view, camera roll and pitch angles}. We also provide \textit{jitter} and \textit{smooth trajectory} to probe flickering predictions between small changes in the viewpoint. See Fig.~\ref{fig:camera} for an overview.

\noindent\textbf{Semantics:} In addition to view changes, we also render images by selecting an object in the scene and changing its occlusion level and scale. In \textit{occlusion} corruption, we generate views of an object occluded by other objects. This is in contrast to random 2D masking of pixels to create an unnatural occlusion effect that is irrespective of image content, e.g. as in~\cite{devries2017improved,naseer2021intriguing}~(See Fig.~\ref{fig:compare2d}). Occlusion rate can be controlled to probe model robustness against occlusion changes. Similarly, in \textit{scale} corruption, we render views of an object with varying distances from the camera location. Note that the corruptions require a mesh with semantic annotations, and are generated automatically, similar to~\cite{armeni20193d}. This is in contrast to~\cite{barbu2019objectnet} which requires tedious manual effort. The objects can be selected by randomly picking a point in the scene or using the semantic annotations.

\begin{figure}[!ht]
\centering
  \includegraphics[scale=0.16]{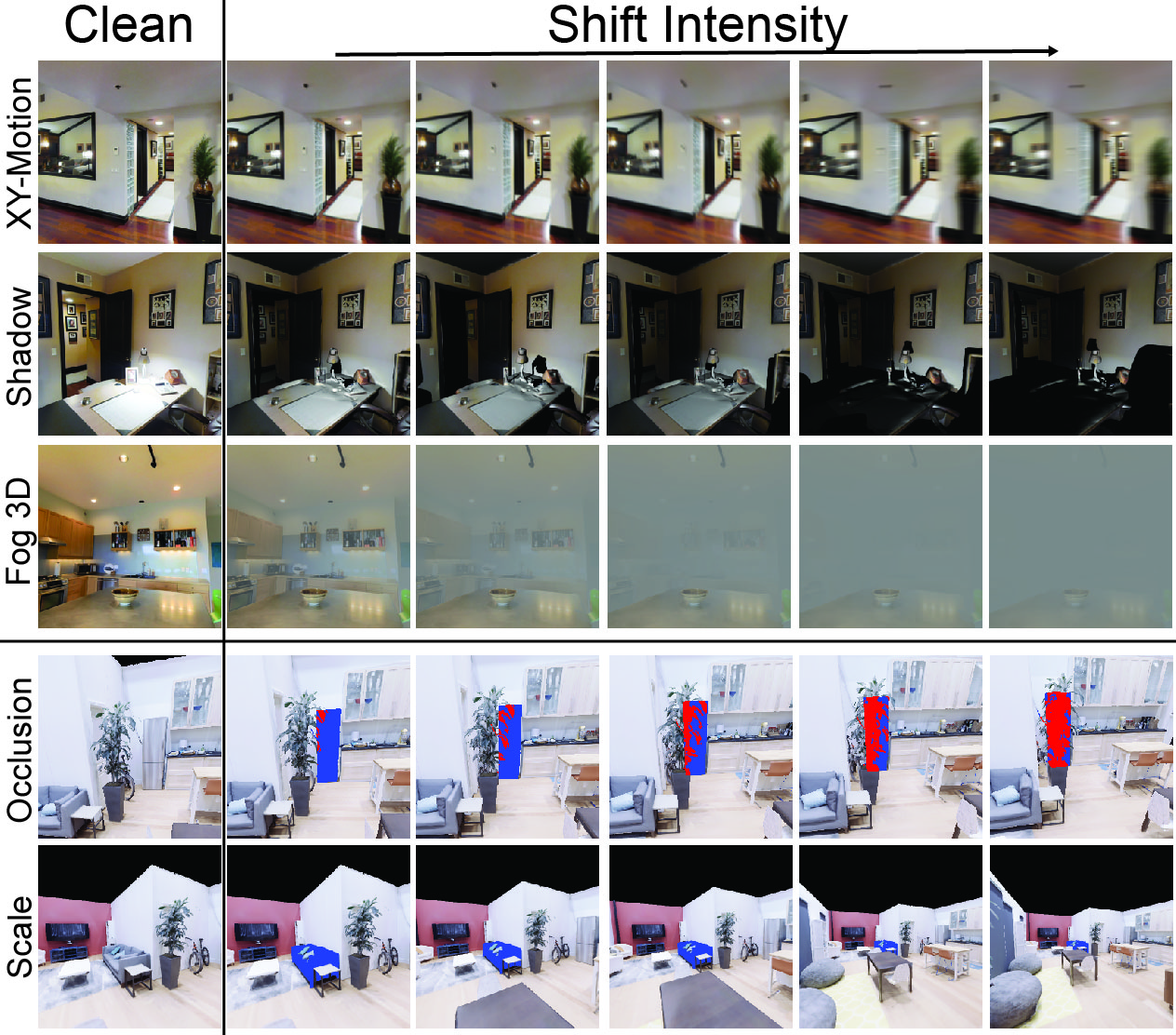}\vspace{-2pt}
%   (Top) Corruptions with increasing severity, leftmost column shows clean images. (Bottom) Camera parameter changes. 
\caption{\footnotesize{\textbf{Visualizations of 3DCC with increasing shift intensities}. \textbf{Top:} Increasing the shift intensity results in larger blur, less illumination, and denser fog. \textbf{Bottom:} The object becomes more occluded or shrinks in size using \textit{calculated viewpoint changes}. The blue mask denotes the amodal visible parts of the fridge/couch, and the red mask is the occluded part. The leftmost column shows the clean images. Visuals for all corruptions for all shift intensities are shown in the \href{https://3dcommoncorruptions.epfl.ch/3DCC_supp.pdf}{supplementary}.}}\label{fig:camera} \vspace{-15pt}
\end{figure}

% , if the dataset includes them.

% the “semantics” paragraph in page 4 need some elaboration. E.g. we don’t say what calculations go into “generating”/“selecting views (basically ray casting). That will confuse some that this process is manual. We should elaborate on this process and emphasize no part of this operation is manual and it’s all coded calculations on mesh+semantic annotations on the mesh. We should also clarify this process needs semantic annotations if one wishes to occlude a certain object in a targeted way (since fig 3 doesn’t include a semantic input modality). Please make sure we drop a citation to 3D scene graph paper somewhere here, since this way of calculating occlusions started from there. Also a citation to objectnet to illustrate a contrast to the tedious manual way of doing a similar operation could go here; that would be effective. Right now you draw constast to the blind 2D occlusion augmentations; it’s a good opportunity to draw the contrast with objectnet style of doing this (and amodal occlusion/detection datasets too)

% to expose model vulnerabilities against scale changes.
% test model against object scale changes.

% \vspace{-5pt}

\noindent\textbf{Noise} corruptions arise from imperfect camera sensors. We introduce new noise corruptions that do not exist in the previous 2DCC benchmark. For \textit{low-light noise}, we decreased the pixel intensities and added Poisson-Gaussian distributed noise to reflect the low-light imaging setting~\cite{foi2008practical}. \textit{ISO noise} also follows a Poisson-Gaussian distribution, with a fixed photon noise (modeled by a Poisson), and varying electronic noise (modeled by a Gaussian). We also included \textit{color quantization} as another corruption that reduces the bit depth of the RGB image. Only this subset of our corruptions is not based on 3D information.

% has a similar distribution where the photon noise is fixed and electronic noise, modeled by Gaussian distribution, increases with higher ISO gains.

% Both noise corruptions are modeled as Poisson-Gaussian distributed~\cite{foi2008practical} where ISO

% We also model low-light and high ISO gain noise corruptions, which do not exist in the original common corruptions to enrich the evaluations on noise corruptions. We model noise as Poisson-Gaussian distributed under low-lighting and high ISO conditions, following~\cite{foi2008practical}. We also included color quantization as another corruption which stems from using a reduced number of bits for the image. 

%  (Bottom) Changes in view e.g. increasing field of view or decreasing camera pitch.
\vspace{-2pt}

\subsection{Starter 3D Common Corruptions Dataset}\label{sec:starterdata}
We release the full open source code of our pipeline, which enables using the implemented corruptions on any dataset. As a starter dataset, we applied the corruptions on 16k Taskonomy~\cite{zamir2018taskonomy} test images.  For all the corruptions except the ones in \textit{view changes} and \textit{semantics} which change the scene, we follow the protocol in 2DCC and define 5 shift intensities, resulting in approximately 1 million corrupted images~($16$k$\times14\times5$). Directly applying the methods to generate corruptions results in uncalibrated shift intensities with respect to 2DCC. Thus, to enable aligned comparison with 2DCC on a more uniform intensity change, we perform a calibration step. For the corruptions with a direct counterpart in 2DCC, e.g. motion blur, we set the corruption level in 3DCC such that for each shift intensity in 2DCC, the average SSIM~\cite{wang2004image} values over all images is the same in both benchmarks. For the corruptions that do not have a counterpart in 2DCC, we adjust the distortion parameters to increase shift intensity while staying in a similar SSIM range as the others. For \textit{view changes} and \textit{semantics}, we render 32k images with smoothly changing parameters, e.g. roll angle, using the Replica~\cite{replica19arxiv} dataset. Figure~\ref{fig:camera} shows example corruptions with different shift intensities. %, changing views, and occlusion levels.

\subsection{Applying 3DCC to standard vision datasets}\label{sec:3dccstandard}

While we employed datasets with full scene geometry information such as Taskonomy~\cite{zamir2018taskonomy}, 3DCC can also be applied to standard datasets without 3D information. We exemplify this on ImageNet~\cite{deng2009imagenet} and COCO~\cite{lin2014microsoft} validation sets by leveraging depth predictions from the MiDaS~\cite{ranftl2019towards} model, a state-of-the-art depth estimator. Figure~\ref{fig:3dfy} shows example images with \textit{near focus}, \textit{far focus}, and \textit{fog 3D} corruptions. Generated images are physically plausible, demonstrating that 3DCC can be used for other datasets by the community to generate a diverse set of image corruptions. In Sec.~\ref{sec:3dfyothers}, we quantitatively demonstrate the effectiveness of using predicted depth to generate 3DCC.

% demonstrating that 3DCC can be used for different downstream tasks by the community to generate a diverse set of image corruptions

% What if your dataset doesn't come with GT depth?

% While we used full scene geometry to generate 3D corrupted data, our framework can also be applied to other datasets such as ImageNet by leveraging depth estimations from state of the art models, e.g. Midas.

% \subsubsection{Using 3DCC as Data Augmentation}
\section{3D Data Augmentation}
While benchmarking uses corrupted images as \textit{test data}, one can also use them as augmentations of \textit{training data} to build invariances towards these corruptions. This is the case for us since, unlike 2DCC, 3DCC is designed to capture corruptions that are more likely to appear in the real world, hence it has a sensible augmentation value as well.  Thus, in addition to benchmarking robustness using 3DCC, our framework can also be viewed as {new data augmentation} strategies that take the 3D scene geometry into account. We augment with the following corruption types in our experiments: \textit{depth of field}, \textit{camera motion}, and \textit{lighting}. The augmentations can be efficiently generated on-the-fly during training using parallel implementations. For example, the depth of field augmentations take $0.87$ seconds (wall clock time) on a single V100 GPU for a batch size of $128$ images with $224 \times 224$ resolution. For comparison, applying 2D defocus blur requires $0.54$ seconds, on average. It is also possible to precompute certain selected parts of the augmentation process, e.g. the illuminations for lighting augmentations, to increase efficiency. We incorporated these mechanisms in our implementation.  %discussed in Sec.~\ref{sec:methods} to generate training data with \textit{near focus}, \textit{far focus}, \textit{XY-motion blur}, \textit{Z-motion blur}, and \textit{flash} corruptions.
We show in Sec.~\ref{sec:3daugs_exp} that these augmentations can significantly improve robustness against real-world distortions. 

\section{Experiments}\label{sec-exp}

We perform evaluations to demonstrate that 3DCC can expose vulnerabilities in models~(Sec.~\ref{sec:3dcc_issues}) that are not captured by 2DCC~(Sec.~\ref{sec:2d3dcc}). The generated corruptions are similar to {expensive realistic synthetic ones}~(Sec.~\ref{sec:3dvsreal}) and are applicable to datasets without 3D information~(Sec.~\ref{sec:3dfyothers}) and for semantic tasks~(Sec.~\ref{sec:3dccsemantic}). Finally, the proposed 3D data augmentation improves robustness qualitatively and quantitatively~(Sec.~\ref{sec:3daugs_exp}). Please see the \href{https://3dcommoncorruptions.epfl.ch}{project page} for a live demo and more extensive qualitative results.

\begin{figure}[!ht]
\centering
  \includegraphics[scale=0.14]{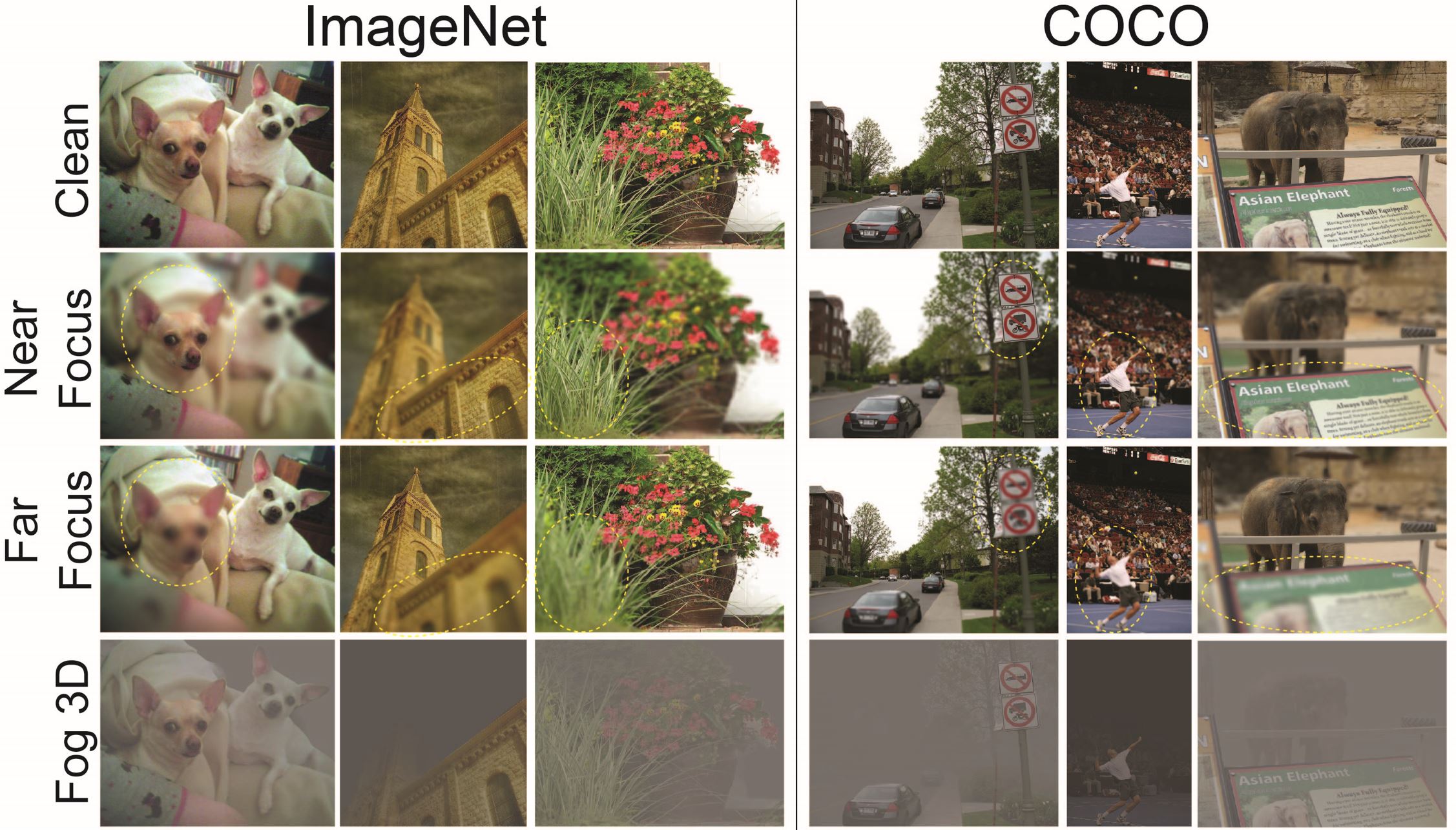}\vspace{-5pt}
%   3D corrupting standard datasets that don't readily come with 3D. TODO: Update caption, mention sota is midas.
\caption{\footnotesize{\textbf{3DCC can be applied to most datasets}, even those that do not come with 3D information. Several query images from the ImageNet~\cite{deng2009imagenet} and COCO~\cite{lin2014microsoft} dataset are shown with \textit{near focus}, \textit{far focus} and \textit{fog 3D} corruptions applied. Notice how the objects in the circled regions go from sharp to blurry depending on the focus region and scene geometry. To get the depth information needed to create these corruptions, predictions from MiDaS~\cite{ranftl2019towards} model is used. This gives a good enough approximation to generate realistic corruptions~(as we will quantify in Sec.~\ref{sec:3dfyothers}).}}\label{fig:3dfy} \vspace{-15pt}
\end{figure}

\subsection{Preliminaries}\label{sec:prelim}

\noindent\textbf{Evaluation Tasks:} 3DCC can be applied to any dataset, irrespective of the target task, e.g. dense regression or low-dimensional classification. Here we mainly experiment with surface normals and depth estimation as target tasks widely employed by the community. We note that the robustness of models solving such tasks is underexplored compared to classification tasks~(See Sec.~\ref{sec:3dccsemantic} for results on panoptic segmentation and object recognition).
% (\textbf{TODO: verify this sentence}) To the best of our knowledge, we are the first to perform a corruption robustness benchmarking on these tasks.
To evaluate robustness, we compute the $\ell_1$ error between predicted and ground truth images.

% Our work is not based on a specific task like ImageNet classification, and it can be used for several tasks thanks to Taskonomy/Omnidata providing labels for a wide set of tasks ranging from low-dimensional classification to dense prediction tasks. For demonstration we consider surface normal and depth estimation tasks as they are widely employed by the community too. Also, to the best of our knowledge, no previous work has done a corruption robustness study on these tasks.

% \subsubsection{Setup}

%\subsubsection{Training Details}

\begin{figure}[!ht]
\centering
  \includegraphics[scale=0.075]{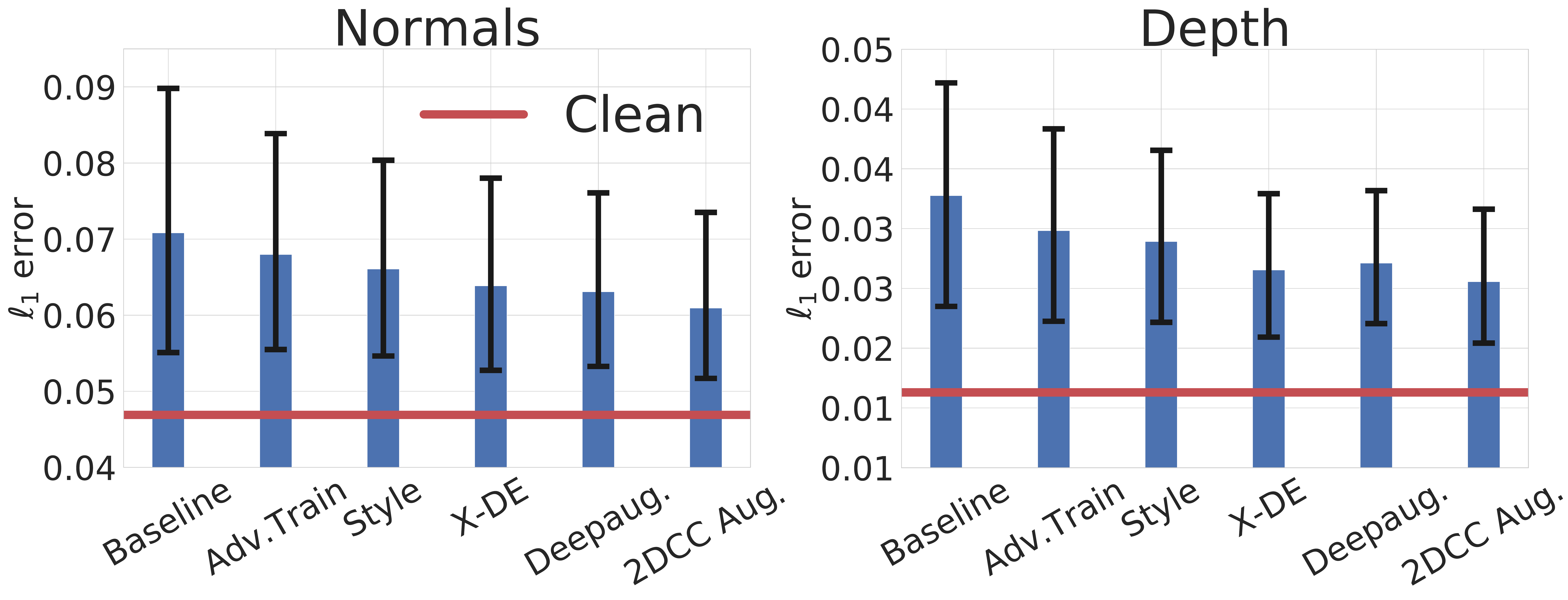}\vspace{-3pt}
%   Comparing different models on 3DCC. 
\caption{\footnotesize{\textbf{Existing robustness mechanisms are found to be insufficient for addressing real-world corruptions approximated by 3DCC.} Performance of models with different robustness mechanisms under 3DCC for surface normals (left) and depth (right) estimation tasks are shown. All models here are UNets and are trained with Taskonomy data. Each bar shows the $\ell_1$ error averaged over all 3DCC corruptions~(lower is better). The black error bars show the error at the lowest and highest shift intensity. The red line denotes the performance of the baseline model on clean (uncorrupted) data. This denotes that existing robustness mechanisms, including those with diverse augmentations, perform poorly under 3DCC.}}\label{fig:comparemodels}
\end{figure}

\noindent\textbf{Training Details:} We train UNet~\cite{ronneberger2015u} and DPT~\cite{ranftl2021vision} models on Taskonomy~\cite{zamir2018taskonomy} using learning rate $5\times 10^{-4}$ and weight decay $2 \times 10^{-6}$. We optimize the likelihood loss with Laplacian prior using AMSGrad~\cite{reddi2019convergence}, following~\cite{Yeo_2021_ICCV}. Unless specified, all the models use the same UNet backbone (e.g. Fig.~\ref{fig:comparemodels}). %Data augmentation models described below are finetuned from the UNet baseline with 50\% clean 50\% augmented data.
We also experiment with DPT models trained on Omnidata~\cite{eftekhar2021omnidata} that mixes a diverse set of training datasets. Following~\cite{eftekhar2021omnidata}, we train with learning rate $1\times 10^{-5}$, weight decay $2 \times 10^{-6}$ with angular~\&~$\ell_1$ losses.

% \ty{Robustness mechanisms evaluated? or Models with robustness mechanisms evaluated.}
% We consider DeepAugment~\cite{hendrycks2021many}, style augmentation~\cite{geirhos2018imagenet}, and adversarial training~\cite{kurakin2016adversarial} as data augmentation strategies that aims to improve robustness.
\noindent\textbf{Robustness mechanisms evaluated:} We evaluate several popular data augmentation strategies: DeepAugment~\cite{hendrycks2021many}, style augmentation~\cite{geirhos2018imagenet}, and adversarial training~\cite{kurakin2016adversarial}. We also include Cross-Domain Ensembles (X-DE)~\cite{Yeo_2021_ICCV} that has been recently shown to improve robustness to corruptions by creating diverse ensemble components via input transformations. We refer to the \href{https://3dcommoncorruptions.epfl.ch/3DCC_supp.pdf}{supplementary} for training details. Finally, we train a model with augmentation with corruptions from 2DCC~\cite{hendrycks2019benchmarking} (2DCC augmentation), and another model with 3D data augmentation on top of that~(2DCC + 3D augmentation).

\subsection{3D Common Corruptions Benchmark}

% \textbf{TODO: Make it clear that we are evaluating robustness over differnet axes (e.g. training data, model arch, aug methods and other robustness mechanisms). And that fig 6 is conditioned on the training data+arch (only uses tasko+unet), fig 8 is conditioned on no augs!!}

% \subsubsection{Performance of models with robustness mechanisms}
\subsubsection{3DCC can expose vulnerabilities} \label{sec:3dcc_issues}
% \ty{4.2.1 Performance of models with robustness mechanisms?}\\

% @Oğuzhan before I forget, let’s make sure we include a unmissable note in the paper (probably at the beginning of the benchmarking using existing networks) with the message that 3DCC benchmark on its own vs the performed benchmarking of the existing models are separate things. The contribution of this paper is the benchmark not the performed analyses. Just like imagenet stands on its own independent of whether the models appleid on it were alexnet, or resnet, or SVM, etc. The models today have certain strengths and weaknesses and tomorrow those will change. It’s the benchmarks that allows us to identify the trends, so one shouldn’t see the value of this paper to provide some experimental study, but the benchmark itself. It’s common sense, but as the authors we should say it once to refresh it in the readers’ mind.

We perform a benchmarking of the existing models against 3DCC to understand their vulnerabilities. However, we note that our main contribution is not the performed analyses but the benchmark itself. The state-of-the-art models may change over time and 3DCC aims to identify the robustness trends, similar to other benchmarks.

% This is because the models may change over time and the   

\noindent\textbf{Effect of robustness mechanisms:} Figure~\ref{fig:comparemodels} shows the average performance of different robustness mechanisms on 3DCC for surface normals and depth estimation tasks. These mechanisms improved the performance over the baseline but are still far from the performance on clean data. This suggests that 3DCC exposes robustness issues and can serve as a challenging testbed for models. The 2DCC augmentation model returns slightly lower $\ell_1$ error, indicating that diverse 2D data augmentation only partially helps against 3D corruptions. 
% some transfer between CC and 3DCC benchmarks exists. 
%  The 2DCC augmentation model yields slightly lower $\ell_1$ errors compared to other models, indicating that diverse 2D data augmentation only partially helps against 3D corruptions.
% The 2DCC augmentation model yields slightly lower $\ell_1$ errors compared to other models, indicating that diverse 2D data augmentation only partially helps against 3D corruptions.

\begin{figure}[!ht]
\centering
  \includegraphics[scale=0.065]{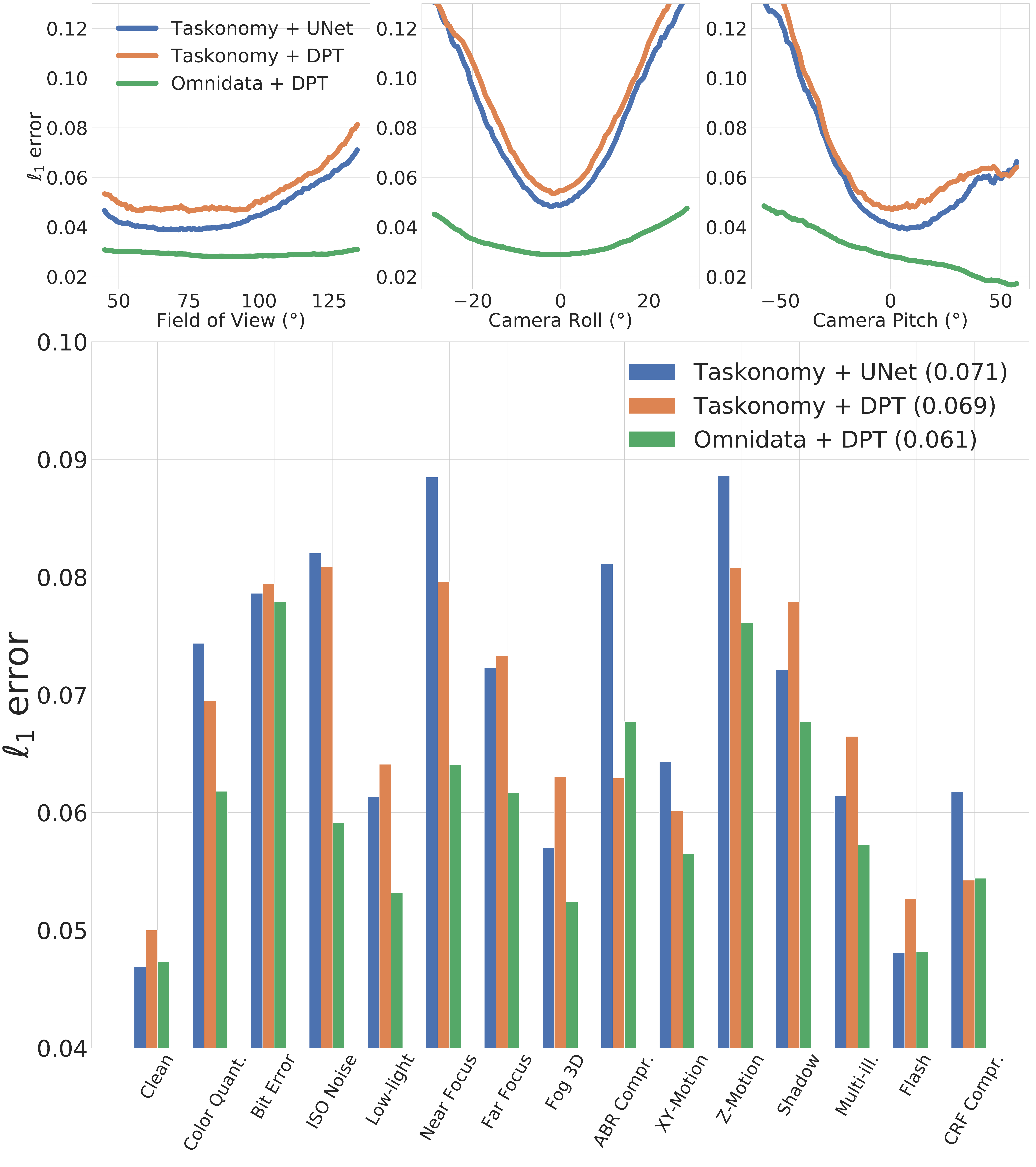}\vspace{-3pt}
%   3D CC exposes robustness issues. Baseline UNet model with no augmentations. Average $\ell_1$ error is reported. Error bars show the performance for minimum and maximum severities. See how much the performance degrades compared to clean input.
\caption{\footnotesize{\textbf{Detailed breakdown of performance on 3DCC.} The benchmark can expose trends and models' sensitivity to a wide range of corruptions. We show this by training models on either Taskonomy~\cite{zamir2018taskonomy} or Omnidata~\cite{eftekhar2021omnidata} and with either a UNet~\cite{ronneberger2015u} or DPT~\cite{ranftl2021vision} architecture. The average $\ell_1$ error over all shift intensities for each corruption is shown~(lower is better). \textbf{Top:} We observe that Taskonomy models are more susceptible to changes in field of view, camera roll, and pitch compared to Omnidata trained model, which is consistent with their methods. \textbf{Bottom:} The numbers in the legend are the average performance over all the corruptions. We can see that all the models are sensitive to 3D corruptions, e.g. \textit{z-motion blur} and \textit{shadow}. Overall, training with large diverse data, e.g. Omnidata, and using DPT is observed to notably improve performance. 
% Thus, the benchmark is an essential tool in diagnosing failure modes of models, revealing their vulnerabilities before they are deployed in the real world.
} }\label{fig:3dcc_issues}
\end{figure}

\noindent\textbf{Effect of dataset and architecture:} We provide a detailed breakdown of performance against 3DCC in Fig.~\ref{fig:3dcc_issues}. We first observe that baseline UNet and DPT models trained on Taskonomy have similar performance, especially on the view change corruptions. By training with larger and more diverse data with Omnidata, the DPT performance improves. Similar observations were made on vision transformers for classification~\cite{dosovitskiy2020image, bhojanapalli2021understanding}. This improvement is notable with view change corruptions, while for the other corruptions, there is a decrease in error from 0.069 to 0.061. This suggests that combining architectural advancements with diverse and large training data can play an important role in robustness against 3DCC. Furthermore, when combined with 3D augmentations, they improve robustness to real-world corruptions~(Sec.~\ref{sec:3daugs_exp}). %For comparison, Omnidata + DPT model yield similar $\ell_1$ error to CC augmentation model, without using corrupted data as augmentations.
\vspace{-12pt}

% We first observe that baseline UNet and DPT models trained on Taskonomy have similar performance. By incorporating larger and more diverse data for training with Omnidata, DPT performance improves. 

\begin{figure}[!ht]
\centering
  \includegraphics[scale=0.033]{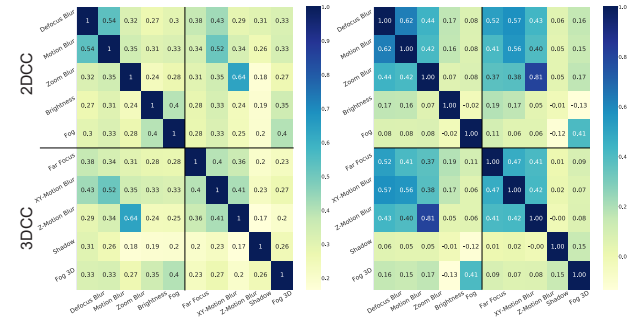}\vspace{-0pt}
%   Error corrs in Normal and RGB domains. Bottom: Corr of residual errors $|clean-corr|$. Which one is more useful? Mean corr for 3d-3d and 2d-3d are lower than 2d-2d indicating 3d is a more diverse benchmark (0.315 vs 0.279 vs 0.296) . Similar conclusions are obtained for depth (see supplementary) 
\caption{\footnotesize{\textbf{Redundancy among corruptions.} We quantified the pairwise similarity of a subset of corruptions from 2DCC and 3DCC by computing their correlations in the $\ell_1$ errors of the surface normals predictions~(left) and RGB images~(right). 3DCC incurs less correlations both intra-benchmark as well as against 2DCC. Thus, 3DCC has a diverse set of corruptions and these corruptions do not have a significant overlap with 2DCC. Using depth as target task yields similar conclusions~(full affinity matrices are provided in the \href{https://3dcommoncorruptions.epfl.ch/3DCC_supp.pdf}{supplementary}). 
% Results on depth and the full affinity (similarity) matrices are given in supplementary.
}}\label{fig:compare2d3dcorrs}
\end{figure}

% \subsubsection{Relation to CC}
% \subsubsection{3DCC vs 2DCC}
\subsubsection{Redundancy of corruptions in 3DCC and 2DCC} \label{sec:2d3dcc}

% The former generates perceptually different and realistic corruptions while the latter does not take scene 3D into account and applies uniform modifications over the image.

%  3DCC incurs less correlation both within the benchmark and with 2DCC compared to 2DCC-2DCC correlations

In Fig.~\ref{fig:compare2d}, a qualitative comparison was made between 3DCC and 2DCC. The former generates more realistic corruptions while the latter does not take scene 3D into account and applies uniform modifications over the image. In Fig.~\ref{fig:compare2d3dcorrs}, we aim to quantify the similarity between 3DCC and 2DCC. On the left of Fig.~\ref{fig:compare2d3dcorrs}, we compute the correlations of $\ell_1$ errors between clean and corrupted predictions made by the baseline model for a subset of corruptions~(full set is in \href{https://3dcommoncorruptions.epfl.ch/3DCC_supp.pdf}{supplementary}). 3DCC incurs less correlations both intra-benchmark as well as against 2DCC~(Mean correlations are $0.32$ for 2DCC-2DCC, $0.28$ for 3DCC-3DCC, and $0.30$ for 2DCC-3DCC). Similar conclusions are obtained for depth estimation~(in the \href{https://3dcommoncorruptions.epfl.ch/3DCC_supp.pdf}{supplementary}). In the right, we provide the same analysis on the RGB domain by computing the $\ell_1$ error between clean and corrupted images, again suggesting that 3DCC yields lower correlations.

\begin{figure}[!ht]
\centering
  \includegraphics[scale=0.065]{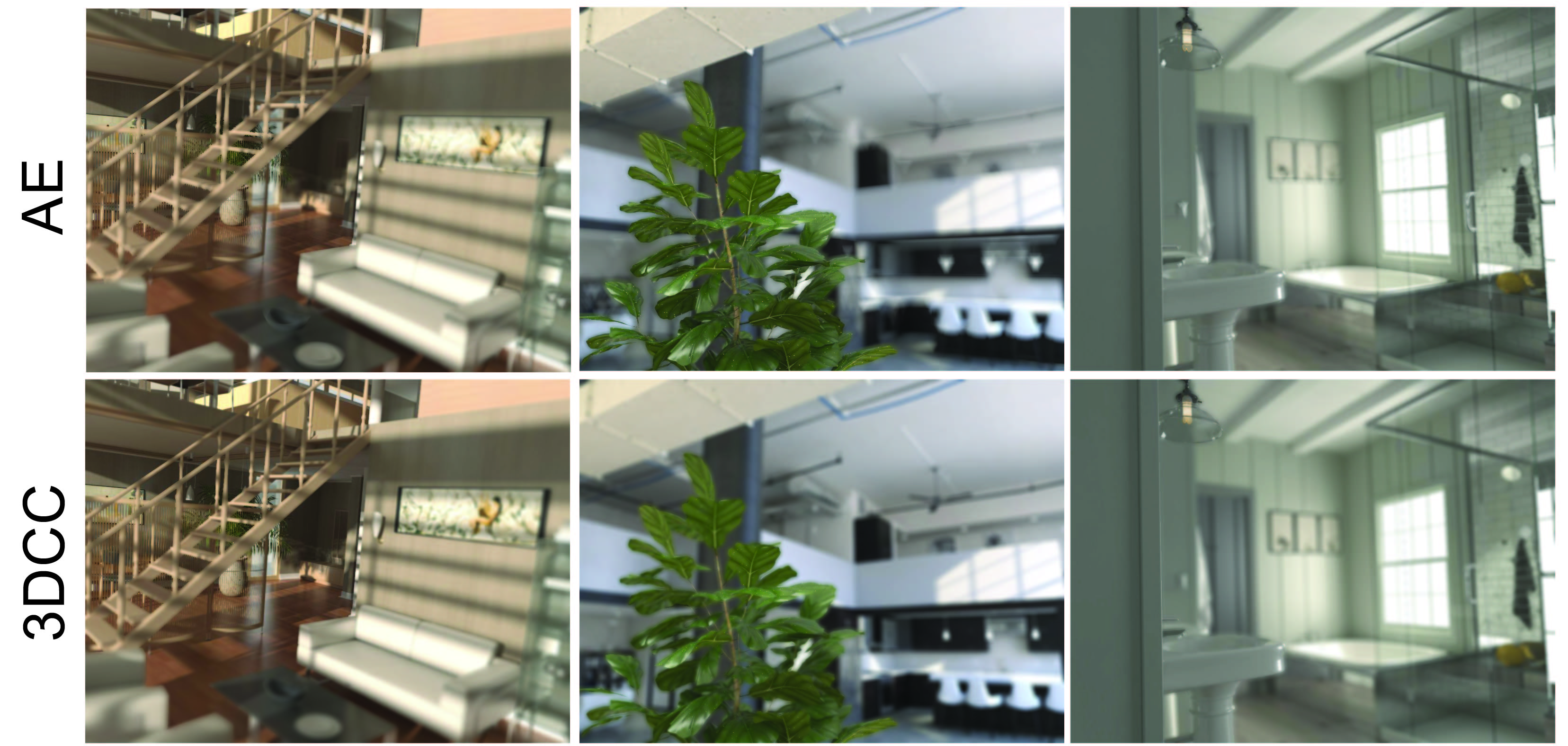}\vspace{-0pt}
%   Compare 3DCC blur with Hypersim AE 
\caption{\footnotesize{{\textbf{Visual comparisons of 3DCC and expensive After Effects (AE) generated depth of field effect} on query images from Hypersim. 3DCC generated corruptions are visually similar to those from AE.}}}\label{fig:compare_ae_3dcc}
\end{figure}
% \vspace{-10pt}
\subsubsection{Soundness: 3DCC vs {Expensive Synthesis}}\label{sec:3dvsreal}

\begin{figure}[!ht]
\centering
  \includegraphics[scale=0.046]{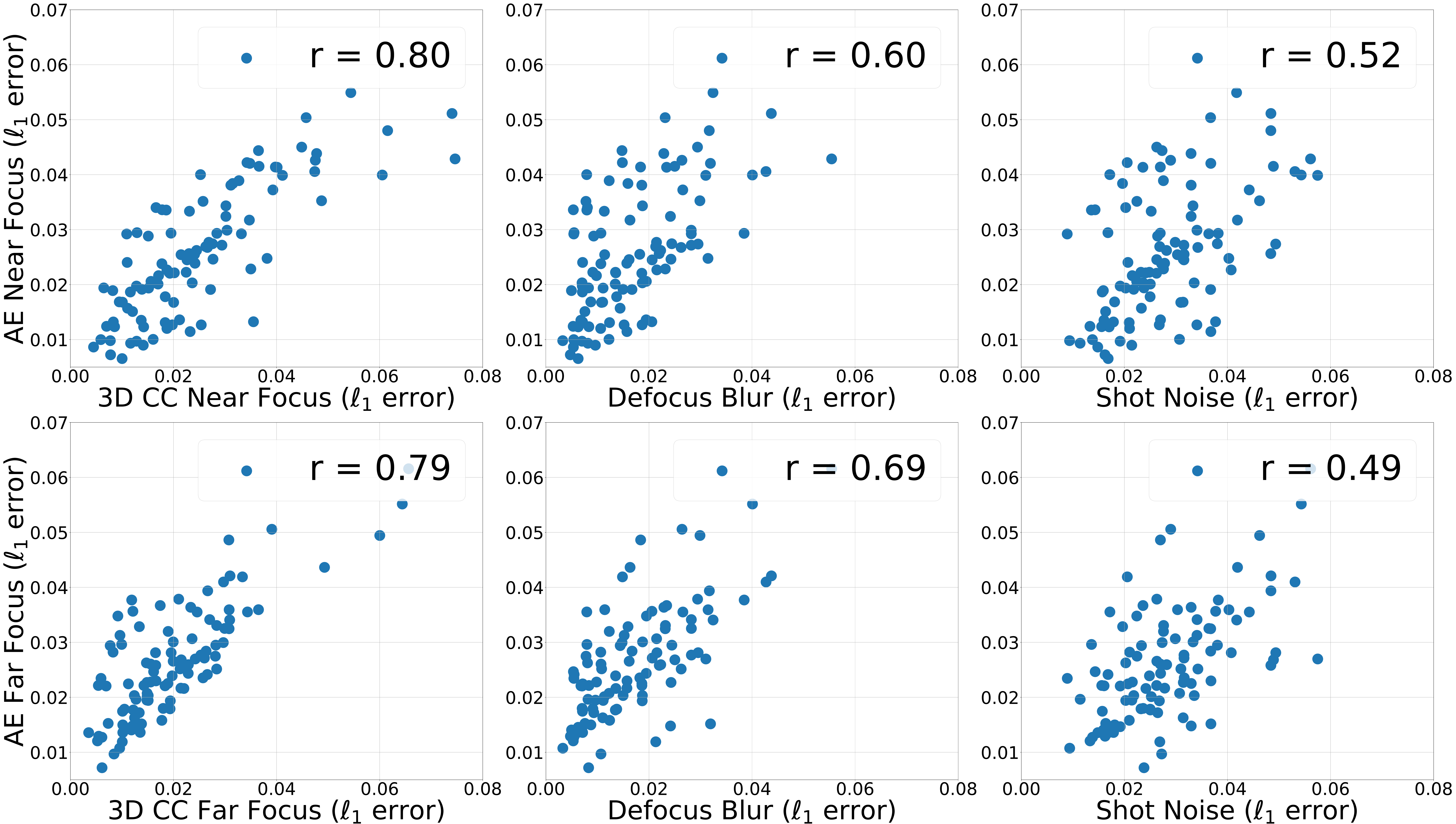}\vspace{-0pt}
%   Trends from 3DCC to expensive synthesis (After Effects - AE). Also compare to 2D corruptions for completeness.(Should we include defocus?)
\caption{\footnotesize{\textbf{Corruptions of 3DCC are similar to {expensive realistic synthetic ones} while being cheaper to generate.} Scatter plots of $\ell_1$ errors from the baseline model predictions on 3DCC against those created by Adobe After Effects~(AE). The correlation between 3DCC near (far) focus and those from AE near (far) focus is the \textit{strongest}~(numbers are in the legend of left column). We also added the most similar corruption from 2DCC (defocus blur) for comparison, yielding weaker correlations~(middle). Shot noise~(right) is a \textit{control baseline}, i.e. a randomly selected corruption, to calibrate the significance of the correlation measure.}}\label{fig:aa_compare}
\end{figure}

3DCC aims to expose a model's vulnerabilities to certain real-world corruptions. This requires the corruptions generated by 3DCC to be similar to real corrupted data. As generating such labeled data is expensive and scarcely available, as a proxy evaluation, we instead compare the realism of 3DCC to synthesis made by Adobe After Effects (AE) which is a commercial product to generate high-quality photorealistic data and often relies on expensive and manual processes. To achieve this, we use the Hypersim~\cite{roberts2021hypersim} dataset that comes with high-resolution z-depth labels. We then generated 200 images that are near- and far-focused using 3DCC and AE.
Figure~\ref{fig:compare_ae_3dcc} shows sample generated images from both approaches that are perceptually similar.
% In supplementary we show that sample generated images from both approaches are similar.
Next, we computed the prediction errors of a baseline normal model when the input is from 3DCC or AE. The scatter plot of $\ell_1$ errors are given in Fig.~\ref{fig:aa_compare} and demonstrates a strong correlation, 0.80, between the two approaches. For calibration and control, we also provide the scatter plots for some corruptions from 2DCC to show the significance of correlations. They have significantly lower correlations with AE, indicating the depth of field effect created via 3DCC matches AE generated data reasonably well.

\subsubsection{Effectiveness of applying 3DCC to other datasets}\label{sec:3dfyothers}

\begin{figure}[h]
\centering
  \includegraphics[scale=0.041]{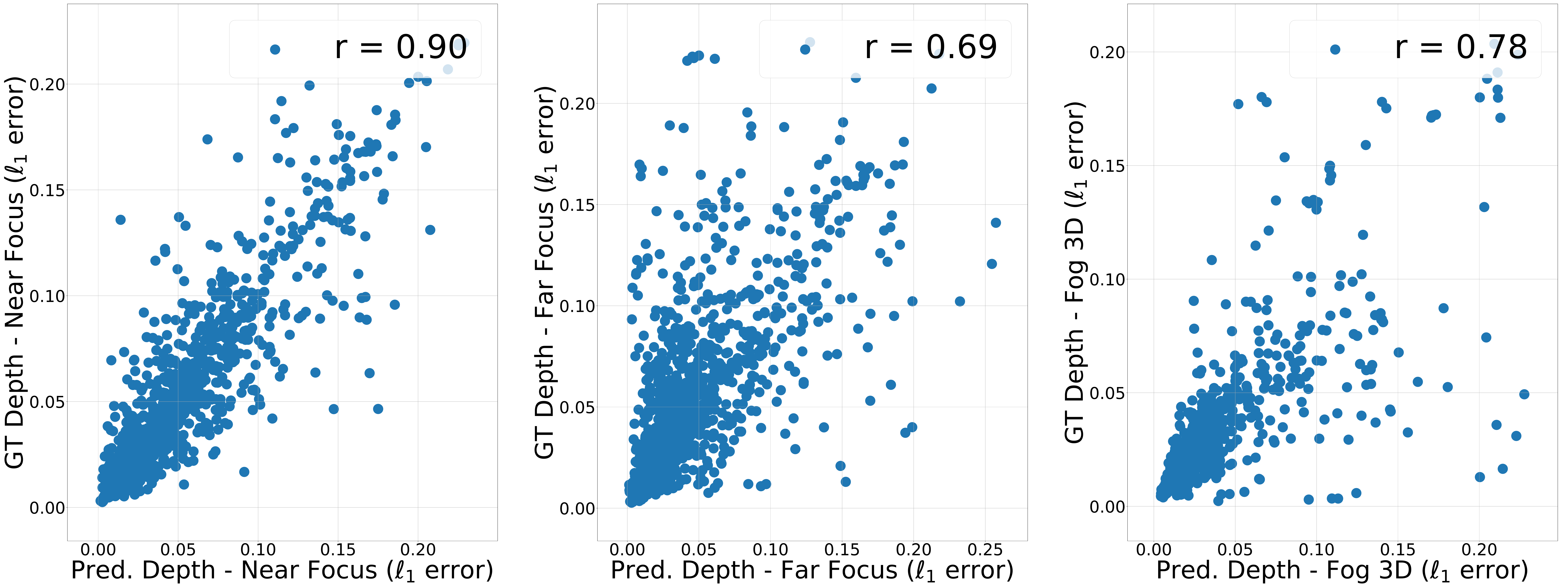}\vspace{-5pt}
\caption{\footnotesize{\textbf{Effectiveness of applying 3DCC without ground truth depth.} 
% {\color{black} Three corruptions from 3DCC are generated using depth predictions from MiDaS~\cite{ranftl2019towards} model (blue) and from ``blind guess depth"~\cite{zamir2020robust} as a control baseline (orange), on unseen Replica data. The plots show the $\ell_1$ errors from the baseline model when corruptions are generated using the predicted depth/blind guess~(x-axis) or the ground truth~(y-axis). The trends are similar when using predicted depth, while using the blind guess results in significantly weaker correlation, thus it is an effective approximation to generate 3DCC~(See the supplementary for more clarifications).}
Three corruptions from 3DCC are generated using depth predictions from MiDaS~\cite{ranftl2019towards} model on unseen Replica data. Scatter plots show the $\ell_1$ errors from the baseline model when corruptions are generated using the predicted depth~(x-axis) or the ground truth~(y-axis). The trends are similar between two corrupted data results, suggesting the predicted depth is an effective approximation to generate 3DCC. See the \href{https://3dcommoncorruptions.epfl.ch/3DCC_supp.pdf}{supplementary} for more tests including control baselines.
% Quantitatively showing that predicted depth can be used to apply 3DCC and it creates similar trends with ground truth depth.
}}\label{fig:3dfy_quant}
\end{figure}

We showed qualitatively in Fig.~\ref{fig:3dfy} that 3DCC can be applied to standard vision datasets like ImageNet~\cite{deng2009imagenet} and COCO~\cite{lin2014microsoft} by leveraging predicted depth from a state-of-the-art model from MiDaS~\cite{ranftl2019towards}. Here, we quantitatively show the impact of using predicted depth instead of ground truth. For this, we use the Replica~\cite{replica19arxiv} dataset that comes with ground truth depth labels. We then generated 1280 corrupted images using ground truth depth and predicted depth from MiDaS~\cite{ranftl2019towards} \textit{without fine-tuning on Replica}. Figure~\ref{fig:3dfy_quant} shows the trends on three corruptions from 3DCC generated using ground truth and predicted depth.
The trends are similar and the correlation of errors is strong~($0.79$). This suggests that the predicted depth can be effectively used to apply 3DCC to other datasets, and the performance is expected to improve with better depth predictions. 

% See the \href{https://3dcommoncorruptions.epfl.ch/3DCC_supp.pdf}{supplementary} for more analysis and quantitative evaluations on ImageNet which suggests that 3DCC can be informative during model development by exposing nonlinear trends and vulnerabilities that are not captured by 2DCC.

%We note that MiDaS model didn't see Replica data during training. 

% Fig.~\ref{fig:3dfy_quant} shows three corruptions from 3DCC generated on 

% As you see in Fig.~\ref{fig:3dfy_quant}, 3DCC can be effectively applied to other datasets that doesn't come with ground truth depth.

% Three corruptions from 3DCC are generated using depth predictions from MiDaS~\cite{ranftl2019towards} model on unseen Replica data. Scatter plots show the $\ell_1$ errors from the baseline model when corruptions are generated using the predicted depth~(x-axis) or the ground truth~(y-axis). The trends are similar between two corrupted data results, suggesting the predicted depth is an effective approximation to generate 3DCC. See the supp. for more tests including control baselines.

\begin{figure*}[!ht]
\centering
  \includegraphics[scale=0.083]{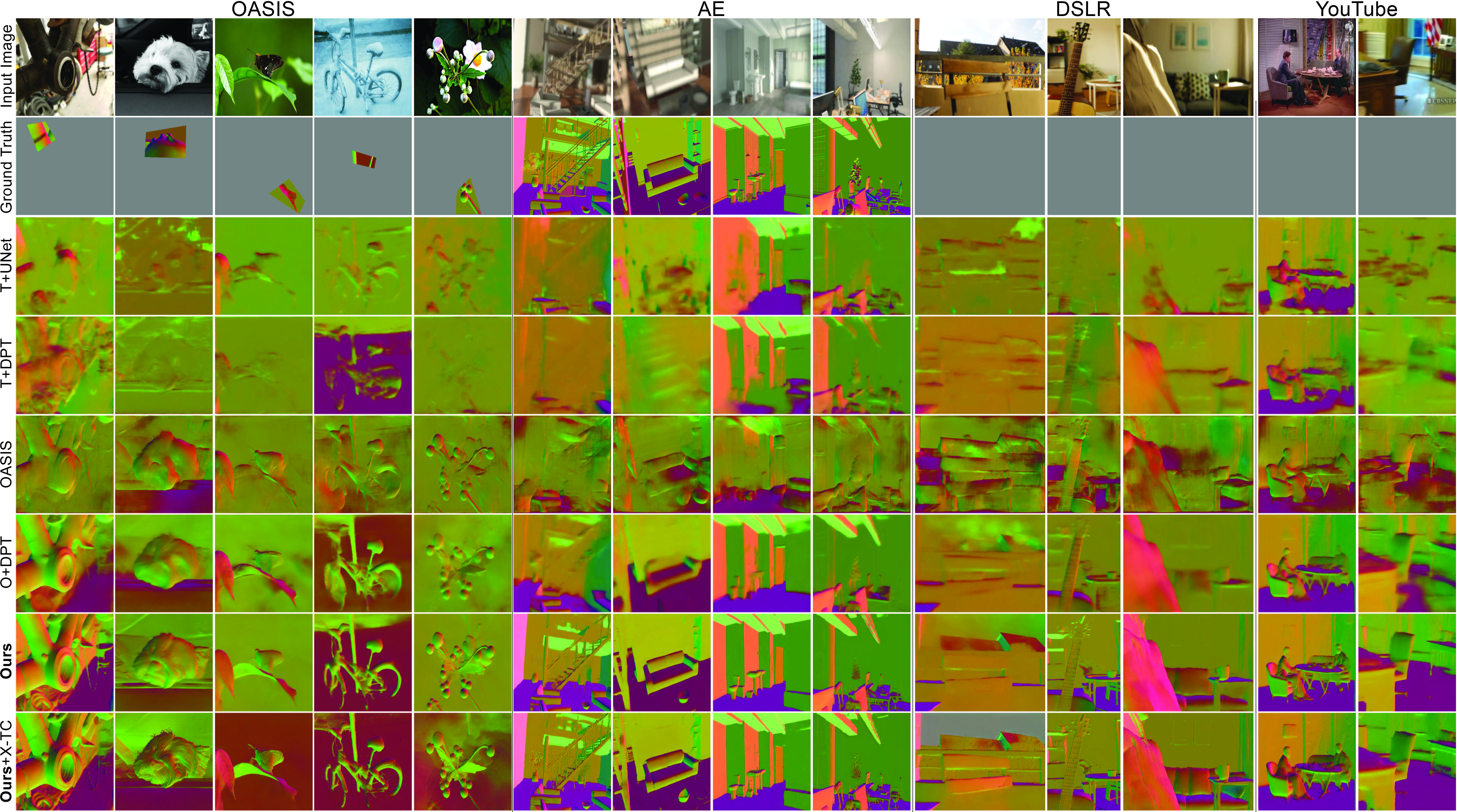}\vspace{-0pt}
\caption{\footnotesize{\textbf{Qualitative results of learning with 3D data augmentation} on random queries from OASIS~\cite{chen2020oasis}, AE~(Sec.~\ref{sec:3dvsreal}), manually collected DSLR data, and in-the-wild YouTube videos for surface normals. The ground truth is gray when it is not available, e.g. for YouTube. The predictions in the last two rows are from the O+DPT+2DCC+3D~(Ours) model. It is further trained with cross-task consistency~(X-TC) constraints~\cite{zamir2020robust}~(Ours+X-TC). They are noticeably sharper and more accurate. See the \href{https://3dcommoncorruptions.epfl.ch}{project page} and \href{https://3dcommoncorruptions.epfl.ch/3DCC_supp.pdf}{supplementary} for more results. A \href{https://3dcommoncorruptions.epfl.ch/\#livedemo}{live demo} for user uploaded images is also available. }}\label{fig:qual}
\end{figure*}

\subsubsection{3DCC evaluations on semantic tasks}\label{sec:3dccsemantic}

The previous benchmarking results were focusing on surface normals and depth estimation tasks. Here we perform a benchmarking on panoptic segmentation and object recognition tasks as additional illustrative 3DCC evaluations. In particular for panoptic segmentation, we use \textit{semantic corruptions} from Sec.~\ref{sec:methods}, and for object classification, we introduce \textit{ImageNet-3DCC} by applying corruptions from 3DCC to ImageNet validation set, similar to 2DCC~\cite{hendrycks2019benchmarking}.

\noindent\textbf{Semantic corruptions:} We evaluate the robustness of two panoptic segmentation models from~\cite{eftekhar2021omnidata} against \textit{occlusion corruption} of 3DCC. The models are trained on Omnidata~\cite{eftekhar2021omnidata} and Taskonomy~\cite{zamir2018taskonomy} datasets with a Detectron~\cite{wu2019detectron2} backbone. See the \href{https://3dcommoncorruptions.epfl.ch/3DCC_supp.pdf}{supplementary} for details.
% The config COCO-PanopticSegmentation/panoptic\_fpn\_R\_50\_3x.yaml was used as a base. The default optimizer was used with a learning rate of 0.02, momentum 0.9 and 500 warmup iterations. The learning rate decays by 10 at iterations 210K and 2.5M. 
% The Detectron was initialised with an ImageNet pre-trained ResNet50 backbone with the first two layers frozen during training.
\begin{figure}[h]
\centering
  \includegraphics[scale=0.1]{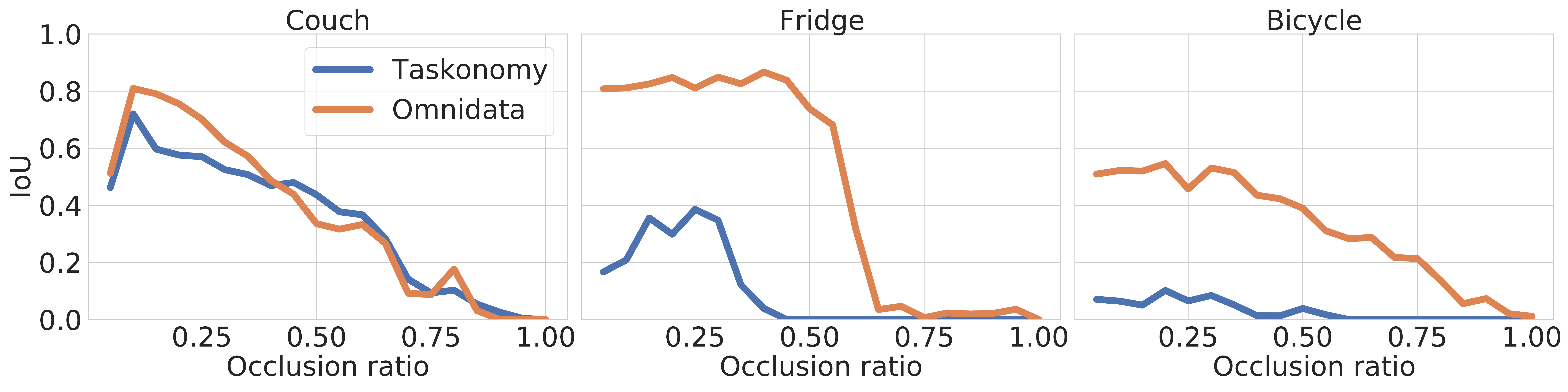}\vspace{-5pt}
\caption{\small \textbf{Robustness against occlusion corruption of 3DCC.} The plot shows the intersection over union (IoU) scores of Detectron models~\cite{wu2019detectron2} for different objects over a range of occlusion ratios. The models are trained on Taskonomy~\cite{zamir2018taskonomy} or Omnidata~\cite{eftekhar2021omnidata} datasets. The occlusion ratio is defined as the number of occluded pixels divided by the sum of occluded and visible pixels of the object. This is computed over the test scenes of Replica~\cite{replica19arxiv}. The plots expose the occlusion handling capabilities of the models and show that the Omnidata trained model is generally more robust than the Taskonomy one. The degradation in model predictions is class-specific and becomes more severe with higher occlusion ratios.}\label{fig:semantic-occ}
\vspace{-12pt}
\end{figure} 

\begin{figure}[!ht]
\centering
  \includegraphics[scale=0.13]{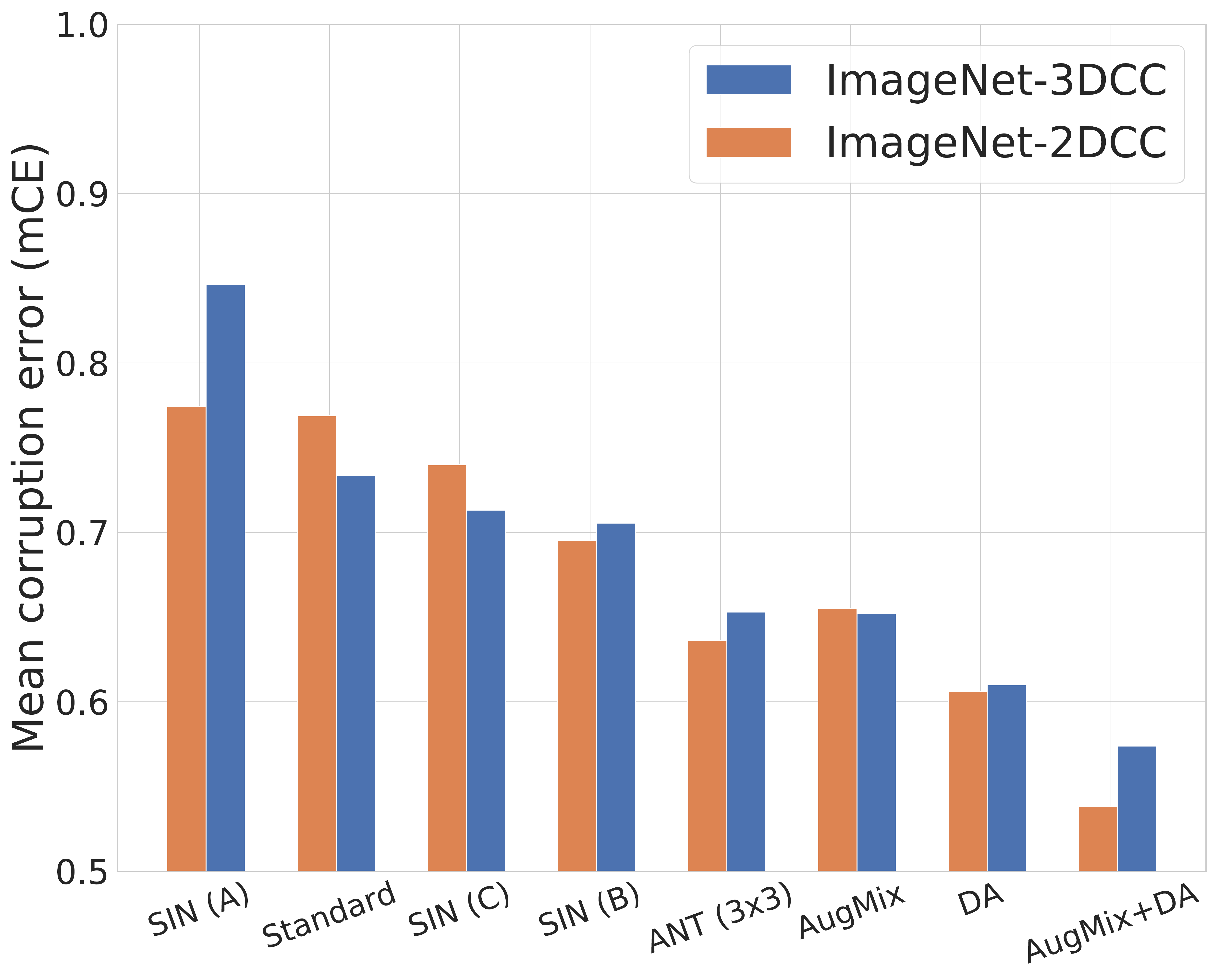}\vspace{-0pt}
%   Comparing different models on 3DCC. 
\caption{\footnotesize{\textbf{Robustness on ImageNet-3DCC and ImageNet-2DCC.}{~Errors on ImageNet validation images corrupted by 3DCC and 2DCC are computed for the models in robustness leaderboards~\cite{hendrycks2019benchmarking,croce2020robustbench}. Following~\cite{hendrycks2019benchmarking}, we compute the mean corruption error~(mCE) relative to AlexNet~\cite{krizhevsky2012imagenet}. The performance degrades significantly against ImageNet-3DCC, thus it can serve as a challenging benchmark. As expected, the general trends are similar between the two benchmarks as 2D and 3D corruptions are not completely disjoint. A similar observation was also made in~\cite{mintun2021interaction} even when the corruptions are \textit{designed to be dissimilar} to 2DCC. Still, there are notable differences that can be informative during model development by exposing trends and vulnerabilities that are not captured by 2DCC, e.g. ANT~\cite{rusak2020simple} has better mCE on 2DCC compared to AugMix~\cite{hendrycks2019augmix}, while they perform similarly on 3DCC. Likewise, combining DeepAugment~\cite{hendrycks2021many} with AugMix improved the performance on 2DCC significantly more than 3DCC. See the \href{https://3dcommoncorruptions.epfl.ch/3DCC_supp.pdf}{supplementary} for more results.
}}}\label{fig:imagenet_3dcc}\vspace{-10pt}
\end{figure}

Figure~\ref{fig:semantic-occ} quantifies the effect of occlusion on the predictions of models, i.e.~how the models' intersection over union~(IoU) scores change with increasing occlusion, for selected objects. This is computed on the test scenes from Replica~\cite{replica19arxiv}. The Omnidata trained model is generally more robust than the Taskonomy one, though we see a decrease in IoU in both models as occlusion increases. The trends are class-specific possibly due to shape of the objects and their scene context, e.g. fridge predictions remain unchanged up until $0.50$ occlusion ratio, while couch predictions degrade more linearly for Omnidata model. This evaluation showcases one use of semantic corruptions in 3DCC, which are notably harder to accomplish using other benchmarks that do not operate based on 3D scans.

\noindent\textbf{ImageNet-3DCC:} We compare performances of the robust ImageNet models~\cite{geirhos2018imagenet,rusak2020simple,hendrycks2019augmix,hendrycks2021many} from RobustBench~\cite{croce2020robustbench} and ImageNet-2DCC~\cite{hendrycks2019benchmarking}~(i.e.~ImageNet-C) leaderboards in Fig.~\ref{fig:imagenet_3dcc}. Following 2DCC, we compute mean corruption error (mCE) by dividing the models errors by AlexNet~\cite{krizhevsky2012imagenet} errors and averaging over corruptions. The performance of models degrade significantly, including those with diverse augmentations. Thus, ImageNet-3DCC can serve as a challenging benchmark for object recognition task. As expected, while the general trends are similar between the two benchmarks as 2D and 3D corruptions are not completely disjoint~\cite{mintun2021interaction}, 3DCC exposes vulnerabilities that are not captured by 2DCC, which can be informative during model development. See \href{https://3dcommoncorruptions.epfl.ch/3DCC_supp.pdf}{supplementary} for further results.

\subsection{3D data augmentation to improve robustness}\label{sec:3daugs_exp}

We demonstrate the effectiveness of the proposed augmentations qualitatively and quantitatively. 
% We train Omnidata DPT models with the proposed 3D+2DCC augmentations using the same training procedure in Sec.~\ref{sec:prelim}. 
% \textbf{Baselines:} 
We evaluate UNet and DPT models trained on Taskonomy~(T+UNet, T+DPT) and DPT trained on Omnidata~(O+DPT) to see the effect of training dataset and model architecture. The training procedure is as described in Sec.~\ref{sec:prelim}. For the other models, we initialize from O+DPT model and train with 2DCC augmentations~(O+DPT+2DCC) and 3D augmentations on top of that~(O+DPT+2DCC+3D), i.e. our proposed model. We also further trained the proposed model using cross-task consistency~(X-TC) constraints from~\cite{zamir2020robust}, denoted as (Ours+X-TC) in the results. Lastly, we evaluated a model trained on OASIS training data from~\cite{chen2020oasis}~(OASIS).

\noindent\textbf{Qualitative evaluations:} We consider \textbf{i.} OASIS validation images~\cite{chen2020oasis}, \textbf{ii.} AE corrupted data from Sec.~\ref{sec:3dvsreal}, \textbf{iii.} manually collected DSLR data, and \textbf{iv.} in-the-wild YouTube videos. Figure~\ref{fig:qual} shows that predictions made by the proposed models are significantly more robust compared to baselines. We also recommend watching the clips and running the live demo on the \href{https://3dcommoncorruptions.epfl.ch}{project page}.

\noindent\textbf{Quantitative evaluations:} In Table~\ref{table-quant}, we compute errors made by the models on 2DCC, 3DCC, AE, and OASIS validation set (no fine-tuning). Again, the proposed models yield lower errors across datasets showing the effectiveness of augmentations. Note that robustness against corrupted data is improved \textit{without sacrificing performance on in-the-wild clean data}, i.e. OASIS.%, is not sacrificed.

\begin{table}[]
\centering
\begin{adjustbox}{width=\columnwidth,center}

\begin{tabular}{|c|c|c|c|c|c|c|c|}
\hline
\backslashbox{Benchmark}{Model} & T+UNet & T+DPT & 
OASIS~\cite{chen2020oasis} & O+DPT & 
% {\begin{tabular}[c]{@{}c@{}}O+DPT+2DCC\\ \textbf{(Ours)}\end{tabular}}
O+DPT+2DCC
% {\begin{tabular}[c]{@{}c@{}}O+DPT+2DCC+3D\\ \textbf{(Ours)}\end{tabular}}

&\textbf{Ours} & \textbf{Ours}+X-TC~\cite{zamir2020robust} \\ \hline\hline
2DCC~\cite{hendrycks2019benchmarking} ($\ell_1$ error)                           & 8.15   & 7.47 &15.31 & 6.43& 5.78  & 5.32 & \textbf{5.29}\\ \hline
3DCC ($\ell_1$ error)                           & 7.08   & 6.89 &15.11 & 6.13& 5.94  & 5.42 & \textbf{5.35}\\ \hline\hline
AE~(Sec.~\ref{sec:3dvsreal}) ($\ell_1$ error)                             & 12.86  & 12.39 & 16.85 & 7.84 & 6.50  & \textbf{4.94} & 5.47\\ \hline
OASIS~\cite{chen2020oasis} (angular error)   & 30.49   & 32.13 &24.63 & 24.42 &  \textbf{23.67}  &  24.65 & 23.89  \\ \hline
\end{tabular}
\end{adjustbox}

\caption{
% \textbf{
% TODO: add results for O+DPT+2D augs (Ours). Im not sure we should bold ``ours" for 2DCC and 3DCC rows because they're on ID data. Revise caption.} Quantitative eval of proposed augmentations on 2DCC, 3DCC, AE, and OASIS without sacrificing clean performance. Full table in supmat.
\textbf{Effectiveness of 3D augmentations quantified using different benchmarks.} $\ell_1$ errors are multiplied by $100$ for readability. The O+DPT+2DCC+3D model is denoted by Ours. We also trained this model using cross-task consistency~(X-TC) constraints from~\cite{zamir2020robust}~(Ours+X-TC). Our models yield lower errors across the benchmarks. 2DCC and 3DCC are applied on the same Taskonomy test images. 
% Note that some of the corruptions in 2DCC and 3DCC overlap with the augmentations used in our models, hence the numbers are reported as a sanity check. 
More results are given in \href{https://3dcommoncorruptions.epfl.ch/3DCC_supp.pdf}{supplementary}. Evaluations on OASIS dataset sometimes show a large variance due to its sparse ground truth. 
}\label{table-quant}
\end{table}

\section{Conclusion and Limitations}

% {\color{brown} We introduce a framework to 1. measure the vulerabilities of models to a wide range of corruptions (see sec...), 2. improve model robustness via augmenting with these corruptions (see sec..).}

We introduce a framework to test and improve model robustness against real-world distribution shifts, particularly those centered around 3D. Experiments demonstrate that the proposed 3D Common Corruptions is a challenging benchmark that exposes model vulnerabilities under real-world plausible corruptions. Furthermore, the proposed data augmentation leads to more robust predictions compared to baselines. We believe this work opens up a promising direction in robustness research by showing the usefulness of 3D corruptions in benchmarking and training. Below we briefly discuss some of the limitations:
% Limitations:
\begin{itemize}[leftmargin=0.0mm,label={}]
\setlength\itemsep{-0.2em}
% \begin{itemize}
    \item \textit{3D quality}: 3DCC is upper-bounded by the quality of 3D data. The current 3DCC is an imperfect but useful \textit{approximation} of real-world 3D corruptions, as we showed. The fidelity is expected to improve with higher resolution sensory data and better depth prediction models. %Hence it's yields
    % Similarly, when 3Dfying other datasets, performance improves as the quality of depth estimation network gets better.
    \item \textit{Non-exhaustive set}: Our set of 3D corruptions and augmentations are \textit{not exhaustive}. They instead serve as a starter set for researchers to experiment with. The framework can be employed to generate more domain-specific distribution shifts with minimal manual effort. 
    % Our set of 3D corruptions are by no means exhaustive, but more like a starter set. Our framework enables researchers to generate new and interesting distribution shifts from training data with minimal manual effort.
    \item \textit{Large-scale evaluation}: While we evaluate some recent robustness approaches in our analyses, our main goal was to show that 3DCC successfully exposes vulnerabilities. Thus, performing a comprehensive robustness analysis is beyond the scope of this work. We encourage researchers to test their models against our corruptions.   
    \item \textit{Balancing the benchmark}: We did not explicitly balance the corruption types in our benchmark, e.g. having the same number of noise and blur distortions. Our work can further benefit from weighting strategies trying to calibrate average performance on corruption benchmarks, such as~\cite{laugros2021using}.
    \item \textit{Use cases of augmentations}: While we focus on robustness, investigating their usefulness on other applications, e.g. self-supervised learning, could be worthwhile.% direction.
    \item \textit{Evaluation tasks}: We experiment with dense regression tasks. However, 3DCC can be applied to different tasks, including classification and other semantic ones. Investigating failure cases of semantic models against, e.g. on smoothly changing occlusion rates for several objects, using our framework could provide useful insights.

\end{itemize}

\vspace{-1pt}
% \noindent\textbf{Acknowledgement:} We thank Zeynep Kar and Abhijeet Jagdev for their help with data generation.
\noindent\textbf{Acknowledgement:} We thank Zeynep Kar and Abhijeet Jagdev. This work was partially supported by the ETH4D and EPFL EssentialTech Centre Humanitarian Action Challenge Grant.

{\small
\bibliographystyle{ieee_fullname}
\bibliography{egbib}
}

\end{document}